\documentclass[preprint,12pt]{elsarticle}  

\usepackage{graphicx}
\graphicspath{{./figures/}{./inkscape/}{../figures/}}
\usepackage{amsmath,amsfonts,amsthm,bm,mathtools} 
\usepackage{amssymb} 
\usepackage{lipsum}
\usepackage{float}
\usepackage{pgfplots}
\usepackage{graphicx,amsmath,upgreek,bm}
\usepackage{subcaption}
\usepackage{siunitx}
\usepackage{makecell}
\usepackage{booktabs}
\pgfplotsset{compat=newest}
\usepackage{multicol}
\usepackage{xcolor,colortbl}
\pgfplotsset{plot coordinates/math parser=false}
\usepackage{multicol} 
\usepackage{amsthm}

\usepackage{geometry}
\usepackage{afterpage} 
\usepackage{longtable}
\usepackage[tableposition=top]{caption}
\usepackage{multirow}
\usepackage[bb=boondox]{mathalpha}
\definecolor{label}{rgb}{0.1,0.25,.65}
\definecolor{title}{rgb}{0.1,0.25,.65}
\definecolor{label}{rgb}{0,0,0}
\definecolor{title}{rgb}{0,0,0}
\definecolor{cite}{rgb}{0.6,0.1,.2}
\usepackage[
font=normalsize,
labelfont={bf,color=label}, 
justification=justified,
format=plain]{caption}

\newlength\fwidth
\usepackage{natbib}
\usepackage{bibentry}
\definecolor{best_acc}{rgb}{0,0.5,0}
\definecolor{refcol}{rgb}{.25,0,1}
\definecolor{mygray}{gray}{0.85}
\definecolor{mycite}{gray}{0.0}
\let\oldbibliography\thebibliography
\renewcommand{\thebibliography}[1]{%
	\oldbibliography{#1}%
	\setlength{\itemsep}{0pt}%
}
\definecolor{mygray}{gray}{0.85}
\definecolor{edit1}{rgb}{.375,.375,.375}
\definecolor{edit2}{rgb}{1,0,0}
\usepackage[misc]{ifsym}
\usepackage{fontawesome}
\usepackage{ragged2e}
\title{ {
        \textbf{
		Machine Learning-Based Ultrasonic Weld Characterization Using Hierarchical Wave Modeling and Diffusion-Driven Distribution Alignment
        }}}

\author[inst1]{\textcolor{title}{Joshua R.~Tempelman}}
 \author[inst2]{\textcolor{title}{Adam J.~Wachtor}}
 \author[inst1]{\textcolor{title}{Eric B.~Flynn}}
 
\affiliation[inst1]{organization={\textcolor{title}{Remote Sensing and Data Science Group,  Los Alamos National Laboratory, Los Alamos NM, USA}
}}
 \affiliation[inst2]{organization={\textcolor{title}{Engineering Institute,  Los Alamos National Laboratory, Los Alamos NM, USA}
 }}

\usepackage{titlesec}
 \titleformat{\section}
{
	\bfseries 
	\large
	\color{label}
}  
{\thesection}  
{1em}  
{}

\titleformat{\subsection}
{
	\bfseries 
	\normalsize
	\color{label}
}  
{\thesubsection}
{1em}
{} 

\usepackage{todonotes}
\usepackage{lineno}

\usepackage{setspace}
\usepackage{mathrsfs}
 
\usepackage[colorlinks=true,
linkcolor=cite,
citecolor=cite,
urlcolor=cite]{hyperref} 
\usepackage{tcolorbox}

\ifdefined\MYCOMMANDS

\else
    \def\MYCOMMANDS{}

    \newcommand{\concat}{\oplus}

    \newcommand{\re}{\text{Re}}
    \newcommand{\im}{\text{Im}}

    \newcommand{\x}{\bm{x}}
    \newcommand{\xEM}{\bar{\bm{x}}}
    \newcommand{\wavefield}{\bm{\varphi}}
    \newcommand{\wavefieldDDPM}{\bm{\varphi}}
    \newcommand{\wavefieldEX}{\bm{\varphi}^{\text{exp}}}
    \newcommand{\nablaEM}{\nabla_{\xEM}}
    \newcommand{\feature}{{\mathsf{F}}}
    \newcommand{\tstar}{{t^{\star}}}

    \newcommand{\reduced}{\mathsf{R}}
    \newcommand{\domain}{\Omega}
    \newcommand{\boundary}{\partial\Omega}
    
    \newcommand{\domainEM}{\Omega^{\reduced}}

    \newcommand{\PML}{\mathsf{PML}}

    \newcommand{\unet}{\mathsf{U}}
    \newcommand{\seg}{{\text{seg}}}
    \newcommand{\inv}{{\text{inv}}}
    
    \newcommand{\EM}{\textsf{EM}}
    \newcommand{\NL}{\textsf{NL}} 
    \newcommand{\dirichlet}{\Gamma_{\mathsf{D}}} 
    \newcommand{\vectorBasis}{\bm{\varXi}}
    \newcommand{\scalarBasis}{\varXi} 
    \newcommand{\ModulationFcn}{\Phi}
\fi 
\begin{document}
	
	\begin{abstract}  
		Automated ultrasonic weld inspection remains a significant challenge in the nondestructive evaluation (NDE) community to factors such as limited training data (due to the complexity of curating experimental specimens or high-fidelity simulations) and environmental volatility of many industrial settings (resulting in the corruption of on-the-fly measurements). Thus, an end-to-end machine learning (ML) workflow for acoustic weld inspection in realistic (i.e., industrial) settings has remained an elusive goal. This work addresses the challenges of data curation and signal corruption by proposing workflow consisting of a reduced-order modeling scheme, diffusion based distribution alignment, and U-Net-based segmentation and inversion. A reduced-order Helmholtz model based on Lamb wave theory is used to generate a comprehensive dataset over varying weld heterogeneity and crack defects.
		The relatively inexpensive low-order solutions provide a robust training dataset for inversion models which are refined through a transfer learning stage using a limited set of full 3D elastodynamic simulations.
		 To handle out-of-distribution (OOD) real-world measurements with varying and unpredictable noise distributions, i.e., Laser Doppler Vibrometry scans, guided diffusion produces in-distribution representations of OOD experimental LDV scans which are subsequently processed by the inversion models.
		 This integrated framework provides an end-to-end solution for automated weld inspection on real data.
	\end{abstract}
	
	\maketitle

	

\section{Introduction}

Welded interfaces are critical in many industries such as energy, aerospace, and infrastructure, and thus the ability to rapid and reliably verify weld integrity remains an important topic of research~\cite{Sun2023}.
Accordingly, a broad range of nondestructive evaluation (NDE) methodologies have been proposed such as visual inspection~\cite{Shafeek2004}, radiography \cite{Kasban2011}, eddy current testing~\cite{Nadzri2018}, acoustic emission testing (AET)~\cite{Saini1998, Grad2004,Zhang2019AE}, and ultrasonic testing (UT)~\cite{Mohandas2024}, each with their unique trade-offs. 
For instance, radiography provides  interpretable images of internal flaws and voids, but  requires long exposure times, is labor intensive, and requires special training to implement~\cite{Alonso2022}.
Moreover, conventional techniques such as visual inspection, liquid penetrant, magnetic particle testing, and infrared thermography can identify surface or volumetric flaws but are either require highly skilled operators  or are costly to implement in-situ \cite{Shafeek2004, Chu2016, Zolfaghari2018, Dorafshan2018, Li2011}. 
AET-based methods provide rapid, low-cost signals of weld instability, yet they suffer from susceptibility to environmental noise and limited ability to quantify defect type, size, or orientation \cite{Roca2007, Gaja2017}. 
In contrast,  UT is easily portable and enables inspection without ionizing radiation, and thus methods such as phased-array ultrasonics and time-of-flight diffraction have been widely adopted~\cite{Bowler2022}.
While UT has proven robust for detecting subsurface cracks and inclusions \cite{Moles2005, Lopez2019}, current inspection practices still require substantial user expertise to configure probes, interpret signals, and validate results.
Thus, automating ultrasonic weld inspection remains a critical need in modern manufacturing and infrastructure maintenance, yet existing approaches face persistent limitations. 
Beyond conventional bulk-wave UT, guided waves have also been studied for weld monitoring, since they can propagate over long distances with low attenuation and, in principle, encode abundant information about weld condition \cite{Moreno2013}. However, their interpretation is complicated by dispersion and scattering at weld geometrical features, and most analysis methods rely on simplifying assumptions such as flat-plate dispersion relations \cite{Kundu2004}. 
This complexity makes guided-wave inspection highly operator-dependent and difficult to scale in practice — but it also highlights an opportunity: because guided waves contain rich signatures of weld defects, they are well-suited to data-driven approaches that can learn to decode these patterns automatically.

Prior research has shown that data-driven methods provide promise for alleviating user involvement \cite{Mirapeix2007,You2014}. 
More recently, deep-learning approaches such as convolutional neural networks (CNNs) and U-Nets \cite{Khumaidi2017, Zhang2019Weld,Munir2018Ultras, Silva2020, Virkkunen2021} have been shown to perform strongly in weld flaw detection and segmentation tasks.
Moreover, these methods are fully automated, and thus AI-driven in-situ monitoring has shown real-time potential for defect detection \cite{Madhvacharyula2022, RocaBarcelo2017}.
However, widespread adoption is limited by two critical bottlenecks: (1) the scarcity of labeled weld defect data, since real flawed specimens are difficult and expensive to curate, and (2) the distribution shift between controlled laboratory scans and noisy industrial inspections \cite{Sun2023, Yuan2024, Tripicchio2020}. While curated datasets such as LoHi-WELD have recently been introduced \cite{BiasuzBlock2024}, the community continues to emphasize the need for scalable synthetic data and robust domain adaptation techniques \cite{Handoko2023}.
These challenges motivate the need for integrated frameworks that combine physics-based reduced-order models, advanced denoising strategies, and deep learning architectures to bridge the gap between simulated and real-world acoustic inspection data.

To address these challenges, recent work has explored physics-based simulation and generative augmentation. Reduced-order models based on the Helmholtz equation and low-order Lamb wave theory can generate large libraries of synthetic UT signals at low computational cost, capturing essential wave–defect interactions without the expense of full elastodynamic finite element analysis \cite{Kundu2004, Koskinen2018}. These proxy datasets can pre-train deep networks before fine-tuning on high-fidelity simulations or experimental weld data \cite{Mohandas2024}. To mitigate domain mismatch, generative methods such as GANs and diffusion models are increasingly applied to NDE. GAN-based augmentation has been shown to improve weld defect classification \cite{Yuan2024, Zhang2025}, while diffusion models offer a more stable approach to denoising and domain adaptation by iteratively refining corrupted signals back to simulation-grade distributions \cite{Ho2020, Yang2024}. This has clear relevance for ultrasonic weld inspection, where environmental volatility and coupling variations corrupt signals in the field. Conditional diffusion schemes can map out-of-distribution industrial scans toward the style of simulation-trained data, enabling ML models to generalize more robustly.

This paper proposes a hybrid solution to automating weld-inspection with guided ultrasonic waves with an emphasis on identifying surface cracks and determining the spatially-resolved weld stiffness.
Training datasets are generated by a hierarchical simulation scheme. First, high fidelity elastodynamic simulations of characteristic welds are used to generated broad but sparsely sampled distribution of various defects, and boundary conditions. 
To overcome the lack of real-world test data and limited availability of expensive high-fidelity simulation data, we develop a reduced-order sequence of Helmholtz models that are based on Lamb wave theory which we term effective medium (EM) solutions---these produce an abundance of training data that closely resemble high-fidelity solutions.
We train separate U-Net modules to predict surface cracks and weld stiffness using steady-state wavefields and their filtered Lamb wave mode components as inputs.
We  numerically validate this approach and confirm that training with EM solutions improves performance on hold-out high-fidelity simulations.
Finally, we propose a diffusion-based distributions alignment scheme to shift out-of-distribution (OOD) experimental measurements into training distribution. Specifically, we generate clean and high-fidelity wavefields based on noisy and OOD experimental measurements by conditioning the reverse diffusion process. We validate this approach using a test weld sample with known defects, and we compare against conventional denoising methods such as denoising CNNs.

The outline of this paper is as follows.
Section~\ref{sec:WaveModel} describes the computational framework for generating both high fidelity and reduced-order solutions,  the datasets designed to train the models, and the experimental data used to ultimately validate the workflow.
Section~\ref{sec:ML} presents the machine learning framework utilized to solve the inverse problem as well as the diffusion-based distribution alignment framework. 
The results of our model are given in section~\ref{sec:results} which details model  performance with respect to the quantify of EM solutions utilized and demonstrates the models effectiveness on simulated high-fidelity solutions and for a set of experimental measurements. 
Lastly, Section~\ref{sec:conclusions} offers concluding remarks and a discussion of possible applications and future work.

\begin{table}[h!]
	\centering
	\caption{Nomenclature}
	\begin{tabular}{ll}
		\hline
		\textbf{Variable} & \textbf{Definition} \\
		\hline
		$A^{(m,n)}$ & Amplitude of Lamb mode $(m,n)$ \\
		$\alpha^{(m,n)}, \beta^{(m,n)}$ & Mode-specific scaling constants \\
		$\bm{f}$ & Force vector \\
		$\Gamma_c$ & Crack boundary \\
		$\Gamma_D$ & Dirichlet boundary \\
		$\Gamma_{wb}$ & Weld–bulk boundary \\
		$J$ & Jacobian of PML mapping \\
		$L_c$ & Crack length \\
		$m$ & Lamb mode symmetry (A = antisymmetric, S = symmetric) \\
		$\mathcal{C}(\xEM)$ & Crack mask function \\
		$c_d$ & Relative crack depth \\
		$d_w$ & Weld depth \\
		$\domain$ & Total computational domain \\
		$\domain_{\PML}$ & Perfectly matched layer (PML) domain \\
		$\boundary$ & External boundaries \\
		$E(\bm{x})$ & Spatially varying Young’s modulus \\
		$\bm\varepsilon$ & Symmetric Eulerian strain tensor \\
		$\bm{x}$ & 3D position vector \\
		$\xEM$ & 2D position vector (reduced domain, $z=h$ plane) \\
		$\nabla$ & Nabla operator \\
		$\nabla_{\xEM}$ & Nabla operator on reduced coordinates \\
		$P_c$ & Probability of crack presence \\
		$\Phi^{(m,n)}(\xEM)$ & Impedance modulation function for mode $(m,n)$ \\
		$\psi^{(m,n)}$ & Scattering mode solution of symmetry $m$, order $n$ \\
		$\psi^{(m,n)}_L$ & Lamb mode solution in $x$–$z$ plane \\
		$\psi_{\EM}(\xEM)$ & Effective medium solution \\
		$\bm\sigma$ & Cauchy stress tensor \\
		$\theta_w$ & Weld angle \\
		$\tilde{\bm{u}}$ & Coordinate-stretched variable in PML \\
		$\bm{u} = [u_x, u_y, u_z]$ & Vector-valued Navier-Lam\'e displacement field \\
		$v_p^{(m,n)}(\xEM)$ & Nominal phase velocity for mode $(m,n)$ \\
		$\bm\varphi$ & Surface field ($u_z$ or $\psi_{EM}$) \\
		$\Omega_b$ & Bulk material domain \\
		$\Omega_w$ & Weld domain \\
		$\Omega^\reduced_c$ & Reduced domain \\
		$\rho$ & Density field \\
		$\lambda, \mu$ & Lamé parameters \\
		$\nu$ & Poisson’s ratio \\
		$\omega$ & Excitation frequency \\
		$x_f$ & Forcing location \\
		\hline
	\end{tabular}
\end{table}
\section{Acoustic Model and Forward Problem}
\label{sec:WaveModel}
Curating a sufficiently large and diverse datasets that faithfully represent the dominant characteristics of the forward problem (i.e., steady-state wave propagation over weldlines) is a necessary precursor for any ML-based inversion task.
This is a challenge for weld inspection~\cite{Cantero2022} since high-fidelity simulations capturing all conceivable phenomena (e.g., anisotropy, nonlinearity, etc) are notoriously expensive and challenging to perform~\cite{Lhemery2000}.
While some researchers have utilized generative techniques to address this~\cite{Virkkunen2021,Yuan2024}, such methods still require a large dataset to train augmentative models or risk biasing generations toward a narrow training distribution, limiting dataset diversity.~\cite{Hu2019}. 
Hence, we propose a hierarchy of forward model complexities as an alternative.
First, a high-fidelity 3D elastodynamic Navier-Lam\'e (NL) model that captures the dominant scattering physics, but at a high computational expense.
Next, a low-order 2D effective medium (EM) model utilizing Lamb wave theory to construct 2D scalar-valued models that capture the dominant wave behavior of the 3D elastodynamic process at a reduced computational cost.
The NL model, being far more expensive to query, serves both to calibrate EM solutions and to provide a low-volume dataset (roughly 10\% of the entire training set) of full fidelity solutions for ML model tuning. 


\subsection{Elastodynamic NL Model}
\label{subsec: NL_Model}
The NL model is constructed as follows.
Let \(\domain = [0,L]\times[0,W]\times[0,H]\in\mathbb{R}^3 \) with external boundary $\boundary$ be the 3D scattering domain with $L,\ W,\ H>0$ being the length, width, and height. Let \(\domain_w\subset\domain\) be a welded region along a 2D path $\gamma(s), s=(x(s),y(s))$  with narrow voids representative of cracks in weldlines, $\Gamma_c\subset\domain_w$.
In some simulations, perfectly matched layers (PMLs) are introduced  along $\boundary$, denoted as $\domain_{\PML}$.
Time-harmonic tractions are applied in $\Gamma_t\subset\partial\domain$ denoted as $\bm{f}(\bm{x})=-F(x,y)\exp(i\omega t) \bm{e}_z$, with $F(x,y)$ being the amplitude profile on the surface $z=h=H/2$, with $\bm{x}=\begin{bmatrix} x&y&z \end{bmatrix}^\intercal$ being the position vector, $\omega$ the frequency,  and $i$ the imaginary unit.
Defining $\domain_b := \domain \setminus \domain_w$ as the bulk material region, the interface \(\Gamma_{wb} := \partial \domain_w \cap \partial \domain_b\) is taken as the boundary between weld and bulk-material such that \( \overline{\domain_w} \cup \overline{\domain_b} = \overline{\domain} \).
We consider constant material properties within $\domain_b$ and spatially varying properties in $\domain_{w}$. 

The elastodynamic force balance in the continuum is governed by the heterogeneous Navier-Lam\'e equations,
\begin{equation}
\begin{aligned}
	\bm{f}(\bm{x})  &= -\rho(\bm{x})\omega^2\bm{u} -\frac{1}{J}\tilde{\nabla}\cdot\left(J\tilde{\bm{\sigma}}(\bm{u})\right) 
	\\
	\tilde{\bm{\sigma}}(\bm{u}) &= 
	\lambda(\bm{x})\text{tr}(\tilde{\bm{\varepsilon}}(\bm{x}))\mathbf{I}+ 2\mu(\bm{x})
	\tilde{\bm{\varepsilon}}(\bm{x})
	\\
	\tilde{\bm{\varepsilon}}(\bm{x}) &= \frac{1}{2}\left(
	\tilde\nabla\bm{u}+ \left(\tilde\nabla\bm{u}\right)^\intercal
	\right)
	\label{EQ:force_balance}
\end{aligned}   
\end{equation}
where $\bm{u}(\bm{x}) = \begin{bmatrix}u_x(\bm{x})&u_y(\bm{x})&u_z(\bm{x}) \end{bmatrix}^\intercal$ is the vector-valued displacement field, 
$\bm{\sigma}(\bm{x}) $ the Cauchy stress tensor,
$\bm{\varepsilon}(\bm{x}) $ the symmetric strain tensor,
and $\rho(\bm{x})$ the density.
$\lambda(\bm{x})$ and $\mu(\bm{x})$ are the spatially-resolved Lam\'e parameters defined by modulus $E(\bm{x})$ and Poisson ration $\nu$
\begin{align}
	\lambda(\bm{x}) = \frac{E(\bm{x}) \nu}{(1 + \nu)(1 - 2\nu)}\\
	\mu(\bm{x}) = \frac{E(\bm{x})}{2(1 + \nu)}
\end{align}
We use the notation $\tilde{\square}$ to denote the complex coordinate stretching in the optionally employed PML boundary layers, e.g.,
$\frac{\partial}{\partial \ell}\mapsto \frac1{\xi_\ell}\frac{\partial}{\partial \ell}, \ \ell = \{x,y,z\}$, 
where $\xi_\ell:\mathbb{R}\to\mathbb{C}$ is the corresponding boundary function,
\begin{equation} 
	\xi_\ell(\bm{x}) = 1+\frac{iP_\ell(\bm{x})}{\omega}, \ \ 
	P_\ell(\bm{x})=
	\begin{cases}
		P_{\max}\bigl(d_\ell(\bm{x}) / w_\PML^{\ell}\bigr)^{a} & \bm{x}\in\domain_{\PML} \\
		0 & \text{otherwise}
	\end{cases}
\end{equation}
with $d_\ell(\bm{x})$ denoting the distance into the PML region and $a=2$ herein.
The Jacobian of the mapping $J = s_x(\bm{x})s_y(\bm{x})s_z(\bm{x})$ is one if $\gamma_{\text{max}}=0$, i.e., if $P_{\text{max}}=0$ or if $\bm{x}\notin\domain_{\PML}$, in which case $\tilde\square=\square$ for all variables and operators.

\begin{figure}\centering
 	\includegraphics[width=\linewidth]{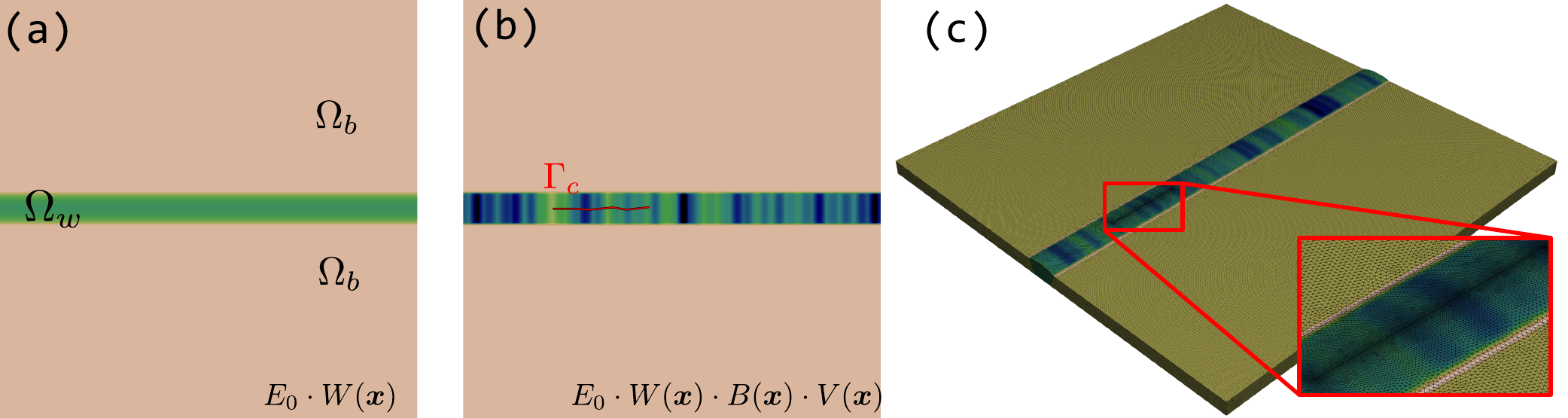}
 	\caption{Weldline stiffness parameterization depicting (a) the nominal weld stiffness profile for a straight weldline through the domain, (b) the modulated weldline, with beading and Gaussian variation applied, and (c) a meshed 3D geometry based on (b).}
 	\label{Fig:GeomEx}
 \end{figure}

The spatial variations of $E(\bm{x})$ are designed to emulate weldline behavior per the baseline material stiffness $E_0$. 
Namely, we consider a composition of several sub-functions:         
 a nominal weldline function $W(\bm{x})$, a bead function ${B}(\bm{x})$ and a pseudo-randomly structured variation function ${V}(\bm{x})$.
$W(\bm{x})$ enforces a baseline stiffness reduction (common in weldlines) about a weld radius $r_w$, ${B}(\bm{x})$ applies a pseudo-periodic beadline profile to stiffness reduction line (Fig~\ref{Fig:GeomEx}(a)), and ${V}(\bm{x})$ takes the form of a smoothly varying Gaussian random field so as to emulate heterogeneity in weldline strength. Altogether, $E(\bm{x})$ is given as,
\begin{equation}
	E(\bm{x}) = E_0\cdot  {W}(\bm{x}) \cdot  {B}(\bm{x})\cdot {V}(\bm{x}).
\end{equation}
The explicit definitions of ${B}(\bm{x})$ and ${V}(\bm{x})$ are detailed in~\ref{APX: Weld_functions}. 
We ensure interface conditions on $\Gamma_{wb}$ delineating the weld region from the bulk material,
\begin{align}
 	[\![ \bm{u} ]\!] &= 0 && \text{(displacement continuity)} \\
 	[\![ \bm{\sigma} \cdot \bm{n} ]\!] &= 0 && \text{(traction continuity)}
\end{align} 
are enforced, with \( [\![ \cdot ]\!] \) denoting the jump condition on $\Gamma_{wb}$.
The 3D geometry incorporates an arc weldline cap which is parameterized as a spline curve based on the weld width and height. 
Moreover, cracks are considered as confined Brownian walks; these are cut from the weldline with a minuscule width, e.g., $\leq 0.01$ inches, (Fig~\ref{Fig:GeomEx}(b)).
 
For NL simulations, we utilize quadratic Lagrange tetrahedral elements using the \texttt{gmsh-4.3.1} API for meshing,  \texttt{FEnICsx-0.6.0} for system matrix assembly, and \texttt{PETSc-3.18.5} as the solver engine.
Further details on the FE implementation of Eq~\eqref{EQ:force_balance} are given in~\ref{APX:FEMS}.
Even with efficient numerical solvers, there is a high memory requirement associated with the 3D vector-valued NL problem on a high-fidelity mesh (Fig~\ref{Fig:GeomEx}(c)).
This is particularly the case for (relatively) high frequency simulations, whereby at least 6 elements per wavelength are required for quadratic elements; this is further exacerbated by the complexity of welded domains as sufficient mesh refinement is required near the weld line and crack to account for the rapidly varying material properties.
For example, a baseline simulation with $W=L=12"$, $H=0.25"$, and $\omega = 2\pi\times225000$ rad/s  necessitates millions of degrees of freedom.
Therefore, a lighter simulation procedure is desired; we describe a proposed alternative the subsequent section.

\subsection{Effective Medium Model}
\label{subsec: EM_Model}
 
We seek a 2D scalar-valued model as a substitute to the 3D NL formulation of section~\ref{subsec: NL_Model} with the goal of capturing the dominant scattering physics.
While plate and shell theories are common substitutes for 3D elasticity, they largely neglect in-plane shear, $\tau_{xy}$. 
For high frequency waves, such as those necessary for resolving fine details pertaining to weld stiffness and continuity, $\tau_{xy}$ may be non-negligible.
Accordingly, plate models overlook important wave scattering patterns of the full elastodynamic model, motivating the need to formulate an alternative approach for the reduced order model.

\begin{figure}[t!]
	\centering
	\begin{subfigure}{\linewidth}
		\includegraphics[width=\linewidth]{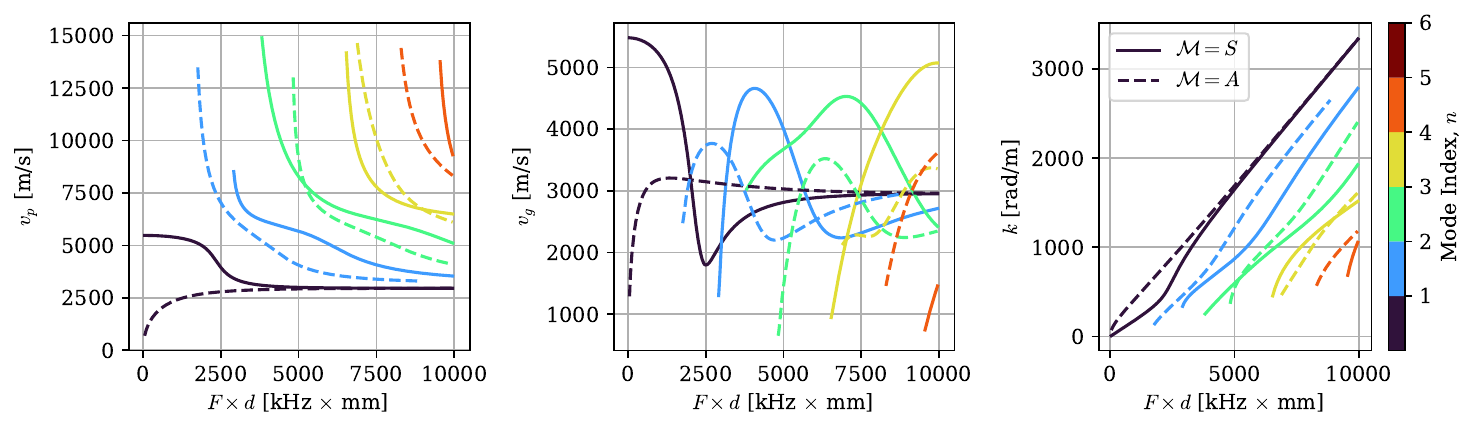}
		\caption{Dispersion Relation}
	\end{subfigure}
	\begin{subfigure}{\linewidth}
		\includegraphics[width=.32\linewidth]{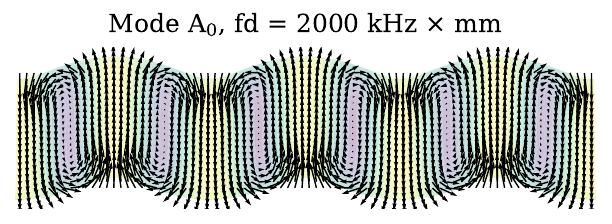}%
		\includegraphics[width=.32\linewidth]{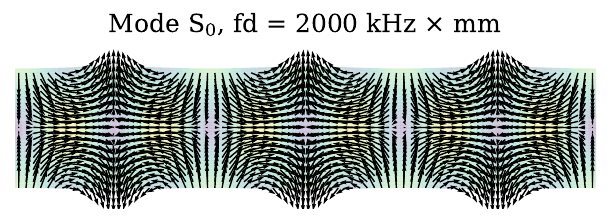}%
		\includegraphics[width=.32\linewidth]{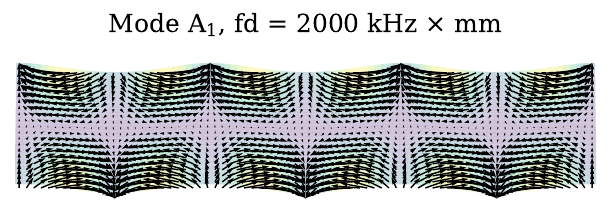}
		\caption{Wave mode solutions}
	\end{subfigure}
	\caption{The solutions of the RL dispersion relation in terms of wavenumber, group velocity, and phase velocity versus the frequency-thickness produce, $F\times d$.}
	\label{Fig:Disperion}
\end{figure}

We consider a 2D scalar-valued Helmholtz-based modeling scheme which we term an \textit{effective medium} (EM) solution.
To capture 3D effects (e.g., coexisting wave modes, traverse and longitudinal shear, etc.), we base our EM solutions on Lamb wave theory. 
The necessary assumptions of Lamb waves are an effectively infinite domain in $x$ and $y$, traction-free surfaces, e.g., $\bm{\sigma}_{zz}=\bm{\sigma}_{xz}=\bm{\sigma}_{yz} = 0$ on $z=\pm h$, and a plane-wave ansatz for the solenoidal and irrotational components of $\bm{u}$ with the wavelength being sufficiently larger than the material thickness.
Sine these assumptions do not restrict material kinematics, the Lamb theory captures the behavior of harmonic elastodynaimcs in $\domain$ under the restriction that $W,L\gg H$ and $\omega$ is selected to satisfy the wavelength assumption.
Hence, focusing our analysis to welded thin-plate structures, we may guide or EM parameterization using the Lamb dispersion relations:
\begin{align}
	\tan(qh)\tan(ph) &= -\frac{\left(k^2-q^2\right)^2}{4k^2pq} \ \ \ & \text{Symmetric}
	\label{Eq:Sym}\\
	\tan(qh)\tan(ph) &= -\frac{4k^2pq}{\left(k^2-q^2\right)^2} \ \ \ & \text{Anti-symmetric}
	\label{Eq:Anti-sym}
\end{align}
Here, $p$ and $q$ are functions of the wavenumber $k$, nominal longitudinal wave speed $c_L =\sqrt{(\lambda_0+2\mu_0)/\rho}$, and nominal traverse wavespeed $c_T=\sqrt{\mu_0/\rho}$,
\begin{equation}
	q = \sqrt{k^2-\frac{\omega^2}{c_L^2}} \ \ \ \ \
	p = \sqrt{k^2-\frac{\omega^2}{c_T^2}}.
\end{equation}
Solutions to Eqs~\eqref{Eq:Sym} and~\eqref{Eq:Anti-sym} produce a denumerable set of propagating wave modes labeled $(m,n)$ herein for symmetry class $m\in[\text{symmetric (S)},\ \text{antisymmetric (A)}]$ and order $n\in\mathbb{Z}_{\geq0}$.
These solutions may be utilized to recover wave propagation characteristics at each frequency-thickness product, as shown by Fig~\ref{Fig:Disperion}(a) depicting the phase velocity, group velocity, and wavenumber of the leading 5 modes for each symmetry.
Moreover, the corresponding displacement fields in the $x$-$z$ plane may be recovered using the wave ansatz, leading to the modes shapes of Fig~\ref{Fig:Disperion}(b)---these may be used to determine the prominence of each mode given a finite surface load. 
A complete discussion regarding the Lamb theory and its solutions can be found in Ref.~\cite{Giurgiutiu2014}.

The phase velocities produced by Eqs~\eqref{Eq:Sym} and~\eqref{Eq:Anti-sym} lead us to a set of 2D PDEs representing  the dominant elastodynamics of the continuum at the surface $z=h$.
Let $\domainEM=[0,H]\times[0,L]\in\mathbb{R}^2$ be a reduced dimensional representation of $\domain$ with corresponding weld domain $\domainEM_{w}\subset\domainEM$ and crack path $\Gamma_c^\reduced\subset\partial\domainEM_w$, and $\xEM := \bm{x}|_{z=h}$ be the 2D position vector at the continuum surface.
The scattering patterns of mode $(m,n)$, denoted $\psi^{(m,n)}$, are described by the Helmholtz model,
\begin{equation}
	\nablaEM \cdot \left( \frac{1}{\xi(\xEM)}
	\left(  
	v_{p}^{(m,n)} \cdot C(\xEM) \cdot \ModulationFcn^{(m,n)}(\xEM)\right)^2
	\nablaEM \psi^{(m,n)}
	\right)
	+ \frac{\omega^2}{\xi(\xEM)} \psi^{(m,n)} = f^{(m,n)},
	\label{EQ:EffectiveMediumPDE}
\end{equation} 
where $v_{p}^{(m,n)}$ is the nominal phase-velocity of the corresponding Lamb mode, $\ModulationFcn^{(m,n)}(\xEM)$ a wavespeed modulation function responsible for parameterizing localized impedance variations, $C(\xEM)$ a crack mask, and $\xi(\xEM)$ the PML coordinate stretching in $\domain_{\PML}^{\reduced}$.
Super-imposing solutions to Eq~\eqref{EQ:EffectiveMediumPDE} for each existing mode of a given frequency-thickness product delivers the cumulative EM solution,
\begin{equation}
	\psi_{\EM}(\xEM;\omega) = \sum_{m,n} A^{(m,n)} \, \psi^{(m,n)}(\xEM)
	\label{EQ:EffectiveMediumSol}
\end{equation}
where $A^{(m,n)}$ is the amplitude of mode $\psi^{(m,n)}(\xEM)$. 
Each $ A^{(m,n)}$ is determined by  projecting $\bm{f}(\xEM)$ onto to the corresponding mode shape, denoted $\bm{\psi}_{L}^{(m,n)}$ (cf.~Fig~\ref{Fig:Disperion}(b).). 
Together with the $z$-polarity of the mode, the projections provide unscaled relative mode prominence,
\begin{equation}
	\tilde{A}^{(m,n)} = 
	 {
	 	\left\|
	 	\int_{\domain} \bm{f}^\dagger(\bm{x})\left( \bm{\psi}_L^{(m,n)}(x,z)\exp\left[i\bm{k}^{(m,n)}\cdot\xEM\right]\right)
	 	 \,  {\rm d}\domain
		\right\|
	}
	\cdot
	{
\left\|
		\frac{
			\int_{-d}^{d} \left|\left(\bm{\psi}_{L}^{(m,n)}(x,z)\right)_z\right| \, {\rm d}z
		}{
			\int_{-d}^{d} \left|\bm{\psi}_{L}^{(m,n)}(x,z)\right| \, {\rm d}z
		} 
		\right\|.
		}
\end{equation}
 The scaled contributions are accordingly taken as $A^{(m,n)}= \tilde{A}^{(m,n)}/\sum_{m,n}\tilde{A}^{(m,n)}$. In the current work, we consider frequencies in the range of 200-250 kHz and nominal plate thicknesses of 0.25 inches. Accordingly, only $A_0$ and $S_0$ need consideration when assembling Eq~\eqref{EQ:EffectiveMediumSol}.

The impedance modulations $\ModulationFcn^{(m,n)}(\xEM)$ for each mode capture the effects of the weld geometry (e.g., the local thickness $H(\x)$), as well as spatially varying modulus $E(\x)$, relative to the nominal plate values $h_0$ and $E_0$,
\begin{equation} 
	\ModulationFcn^{(m,n)}(\xEM) = \left(
	\frac{E(\xEM) H(\xEM)^{\alpha^{(m,n)}}}{E_0h_0}\right)^{\beta^{(m,n)}},
\end{equation} 
where $\alpha^{(m,n)}$ and $\beta^{(m,n)}$ are mode-specific scaling constants. The selected values utilized in this study are given in Table~\ref{Table:Scaling}; these were first selected based on qualtivative assement of $\psi_L^{(m,n)}$ and tuned ad-hoc by calibrating against NL solutions between 200 and 400 kHz.
The crack mask $C(\xEM)$ is given a smoothed binary mask, with the binary mask $m(\xEM)$ taking the value $1$ or $1-c_d^2$, with $c_d$ being the  relative crack depth for the corresponding 3D geometry,
\begin{equation} 
	C(\xEM) =\left(\mathcal{G}_\sigma\ast m\right)(\xEM), \ \ \ \ m(\xEM) = \begin{cases}
		1-c_d^2 & \xEM\in\Gamma^\reduced_c\ \\
		1 & \text{otherwise}.
	\end{cases}
	\label{EQ:CrackMask}
\end{equation} 
$\mathcal{G}_{\sigma}$ denotes a Gaussian kernel of bandwidth $\sigma$ and $\ast$ is convolution---this smoothing ensures numerical stability for the FE solutions. 
The quantity $c_d$ is squared in Eq~\eqref{EQ:CrackMask} to emphasize the effect of shallower cracks, i.e., $c_d<0.5$, which were empirically found to produce disproportionately larger wavefield disturbances relative to their depth in NL solutions compared to EM.
\begin{figure}[t!]\centering
	\begin{subfigure}{\linewidth}
		\includegraphics[width=\linewidth]{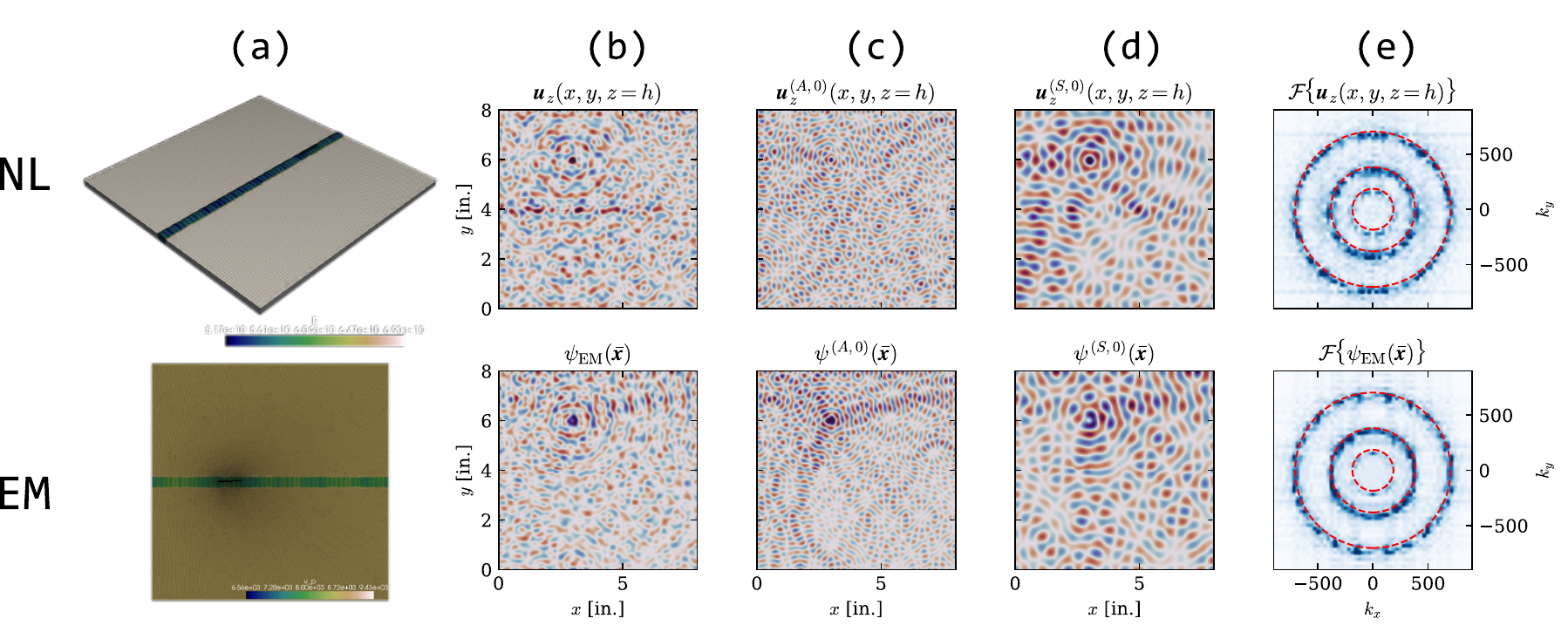}
	\end{subfigure}
	\caption{Confirmation of the Lamb-wave assumptions. (a) A full D continuum for computing NL solutions and reduced 2D continuum for generating EM solutions. (b) The resulting out-of-plane displacement fields at $z=h$ with their filtered (c) $(A,0)$ and (d) $(S,0)$ components showing qualitative agreements. (e) The spectra of each wavefield show aggregated energy along pre-computed Lamb wave mode numbers (dashed-lines), confirming the alignment of the high-fidelity problem with Lamb theory. }
	\label{Fig:ModelComp}
\end{figure}

Solutions to Eq~\eqref{EQ:EffectiveMediumPDE} were computed with quadratic Lagrange elements on a triangular  mesh. Local refinement was implemented near cracks if they exist in the domain, and the computational engines for constructing the mesh, system matrices, and solutions follow from section~\ref{subsec: NL_Model}.
Further details on the FE implementation of EM solutions are given in~\ref{APX:FEMS}.

\subsection{Model Comparisons}
\label{subsec: EM_NL_Comparison}
%
Figure~\ref{Fig:ModelComp} depicts the comparison for a given wavefield simulation at 350 kHz for a 0.25-inch steel plate with a center weldline. 
The  solution field at the surface, denoted $\bm{\varphi}$ herein corresponding to $\bm{u}_z(\xEM)$ for NL or $\psi_{\EM}(\xEM)$  for EM solutions, are given by
Fig~\ref{Fig:ModelComp}(b). The filtered $(A,0)$ and $(S,0)$ components given by \ref{Fig:ModelComp}(c) and \ref{Fig:ModelComp}(d), respectively, which show a qualitative agreement in the solution and filtered fields between the full-order and reduced-order models---these are computed by applying radial Gaussian filters in the wavenumber domain centered at the predicted spatial frequencies from Lamb dispersion. Namely,
\begin{equation}
	\wavefield^{(m,n)} =\mathcal{F}^{-1}_{{\bm{k}}} \left\{ \mathcal{F}_{\xEM}\left\{ \wavefield\right\} 
	\exp\left[-\frac{\left(\bm{k}-{k}^{(m,n)}\right)^2}{2\left(\sigma^{(m,n)}\right)^2}
	\right]\right\}
\end{equation}
where $\bm{k}$ is the wavevector, $k^{(m,n)}$ the wavenumber of mode $(m,n)$, and with
$\mathcal{F}^{-1}_{{\bm{k}}}$ and $\mathcal{F}_{\xEM}$ denoting the inverse and forward 2D spatial Fourier Transform, respectively. The bandwidth $\sigma^{(m,n)}$ is selected such that a half-power decay is achieved midway between adjacent center wavenumbers.
Lastly, the spectrum of each are given by Fig~\ref{Fig:ModelComp}(e), showing that the same dominant scattering frequencies used to generate $\psi_{\EM}(\xEM)$ (denoted with dashed red lines) dominate $\bm{u}_z(\xEM)$.

\begin{table}\centering
	\caption{Scaling factors for $\ModulationFcn^{(m,n)}(\xEM)$}
	\label{Table:Scaling}
	\begin{tabular}{llll}\hline
		& (A,0) & (S,0) & (A,1)\\\hline
		$\alpha^{(m,n)}$ &1.1&1.3&1.0\\
		$\beta^{(m,n)}$  &0.5&0.5&0.5\\\hline
	\end{tabular}
\end{table}

\subsection{Problem Domains}
\label{subsec:domains}
\begin{figure}[t!]
	\includegraphics[width=\linewidth]{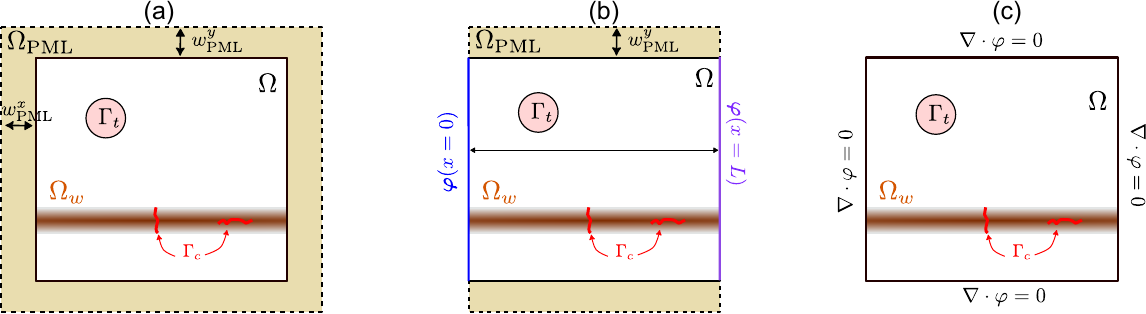}
	\caption{Problem Domains for the (a) scattering, (b) cylinder, and (c) coupon problem classes with $w_{\PML}^s$ denoting the PML boundary width in the $s=\{x,y\}$ directions.
		The variable $\wavefield$ denotes either $\bm{u}(\bm{x})$ or $\psi^{(m,n)}(\xEM)$ for elastodynamic and EM solutions, respectively. 
	}
	\label{Fig:Probs}
\end{figure}

Three classes of boundary conditions are considered representing three prominent weld inspection tasks.
First, a scattering framework for which exterior boundaries are enclosed by $\domain_{\PML}$, e.g., $w_{\PML}>0$ in both $x$ and $y$ leading to a PML-PML configuration.
This emulates inspection scenarios whereby the extent in $x$ and $y$ is effectively infinite and thus boundary reflections are negligible, representative of numerous NDE inspection tasks, i.e., aerospace fuselages and large storage tanks.
Second, a semi-periodic domain with periodic boundaries enforced in $x$ and PML boundaries in $y$, leading to a periodic-PML configuration.
This emulates a semi-infinite cylinder indicative of pipe inspection problems.
Lastly, an open-boundary problem with no PML attenuation or periodicity, leading to a free-free configuration. This emulates the measurement of relatively small coupon samples or small manufactured components.
The corresponding boundary conditions are depicted by Fig~\ref{Fig:Probs} and are described as follows,
\begin{align}
	&\textsf{Scattering:}\ \	&{\nabla}\cdot\wavefield=0 \ \text{on} \ \partial\domain, \ \ \  w_{\PML}^{x}=\chi, \ \ w_{\PML}^{y}=\chi  
	\label{EQ:BC_scatter}
	\\
	&\textsf{Periodic:}\ \	&\wavefield(x=0)=\wavefield(x=L), \ \ {\nabla}\cdot\wavefield = 0 \ \text{on} \ y=0,L, \ \ \ w_{\PML}^{x}=0, \ \ w_{\PML}^{y}=\chi 
	\label{EQ:BC_pipe}
	\\
	&\textsf{Free-Free:}\ \		&{\nabla}\cdot\wavefield=0 \ \text{on} \ \partial\domain, \ \ \ w_{\PML}^{x}=0, \ \ w_{\PML}^{y}=0
	\label{EQ:BC_coupon}
\end{align}
where $w_{\PML}^s$ is the depth of the PML boundary in the $s$-coordinate direction,  $\chi$ is the nominal depth selected to be at least 3 wavelengths of the highest wavenumber mode, and again $\wavefield$ denotes either $\bm{u}(\bm{x})$ or $\psi^{(m,n)}(\xEM)$ for elastodynamic and EM solutions, respectively.
The domain dimensions of the scattering and periodic classes are nominally taken to be 8"$\times$8"$\times$0.25", whereas the coupon (free-free) problem class is considered on a 4"$\times$8"$\times$0.25" domain to reflect available experimental samples.
All elastodynamic simulations enforce Neumann boundaries on $z=\pm h$.

\subsection{Dataset Design} 
\label{subsec:Dataset_Design}

Simulation datasets were generated to ensure sufficient diversity of key domain parameters in both training and testing.
Namely, the weld width $L_w$, weld angle $\theta_w$, force location $\x_{f}$,
nominal weld stiffness reduction $W_0$, and relative weld variation amplitude $V_0$, are drawn form normal distributions---this emulates deviations from a nominal welding and inspection processes due to natural process variability from  nominal welding and measurement schemes.
The crack length $L_c$ and crack depth $d_c$ are both drawn from uniform distributions---this emulated deviations in crack nature which are not suspected to be biased toward any nominal length or depth.
Finally, we consider problem domains both with and without cracks, with the probability of a crack occurring denoted as $\mathcal{P}_c$ being sampled from a Bernoulli---this prevents biasing the proposed data-driven inversion framework from always predicting cracks when none are present.
These parameters are varied per the distributions defined in Table~\ref{Table:Dataset} for each problem domain discussed in section~\ref{subsec:domains} with the exception of $\theta_w$ which is fixed at 0 for the periodic pipe problem to ensure continuity for the periodic boundary conditions.
The distributions of Table~\ref{Table:Dataset}  ensure that the models are trained on a diversity of meta-parameters emulating real-world variations. 
For instance, scanning fixtures such as scanning laser doppler vibrometers (LDVs) may not always be perfectly oriented, and hence the weld may appear shifted or tilted relative to the scan. 
Moreover, the placement of transducers is subject to change, so $\bm{x}_f$ in the simulated models should emulate this. 
Lastly, weldlines may vary in width and relative strength, and these nominal parameters must be diversely sampled as well to mitigate bias in training. 

\begin{table}[t!]\centering
	\caption{Distribution definitions for dataset design variables with $\mathcal{N}$, $\mathcal{U}$, and $\mathcal{B}$ denoting normal, uniform, and Bernoulli distributions.}
	\label{Table:Dataset}
	\begin{tabular}{lll}\hline
		Parameter 			&Description	& Distribution\\\hline
		$L_c$ 			&  Crack Length		&$\mathcal{U}(L/50,L/2)$		\\
		$d_c$ 	&Depth of Crack 		&$\mathcal{U}(H/10,H)$				\\
		$\mathcal{P}_c$ 	& Probability of Crack 		&$\mathcal{B}(0.5)$				\\
		$\theta_w$ 			& Weld Angle	 	&$\mathcal{N}(0,\pi/4)$		\\
		$d_w$ 			   &	Weld Depth 	&$\mathcal{N}(H/5,H/20)$	\\ 
		$\bm{x}_f$ 			   &	Force Location &$\mathcal{N}(\bm{x}_{f_0},\text{diag}([W/4,H/6]))$	\\ 
		\hline
	\end{tabular}
\end{table}

The simulation frameworks of sections~\ref{subsec: NL_Model} and~\ref{subsec: EM_Model} are utilized to curate datasets by sampling the distributions of simulation parameters.
We consider a data enrichment scheme whereby a majority of generated examples (10,000 per problem class) are provided by solving Eq~\eqref{EQ:EffectiveMediumSol}, with a much smaller set of full elastodynamic simulations (1,000 per problem class) provided by solving Eq~\eqref{EQ:force_balance}.
The densely sampled distribution of EM solutions serve as an enrichment dataset to pre-train inversion and generative models on, before then tuning on them to the limited samples of NL solutions, as explained in subsequent sections.

\subsection{Experimental Measurements}
\label{subsec:ExMeas}
\begin{figure}\centering
	\begin{subfigure}{.5\linewidth} 
		\includegraphics[height=.5\linewidth]{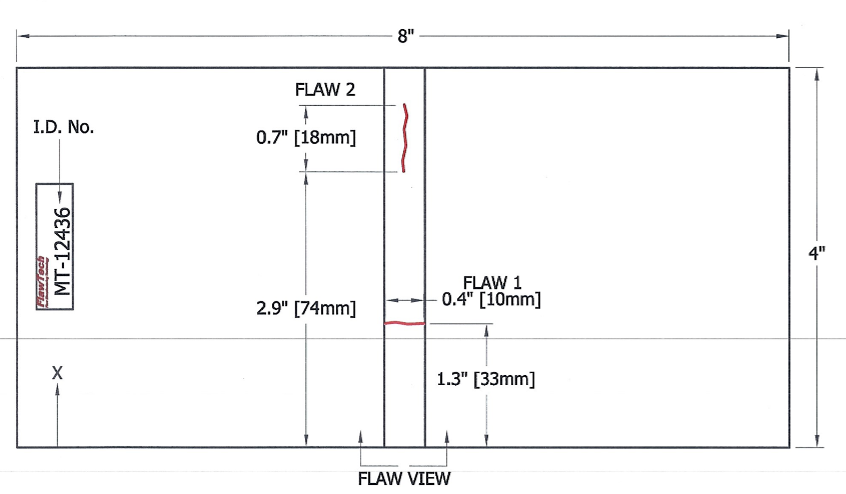}
		\caption{Experimental Weld Sample}
		\end{subfigure}%
		\begin{subfigure}{.5\linewidth}
			\includegraphics[height=.5\linewidth]{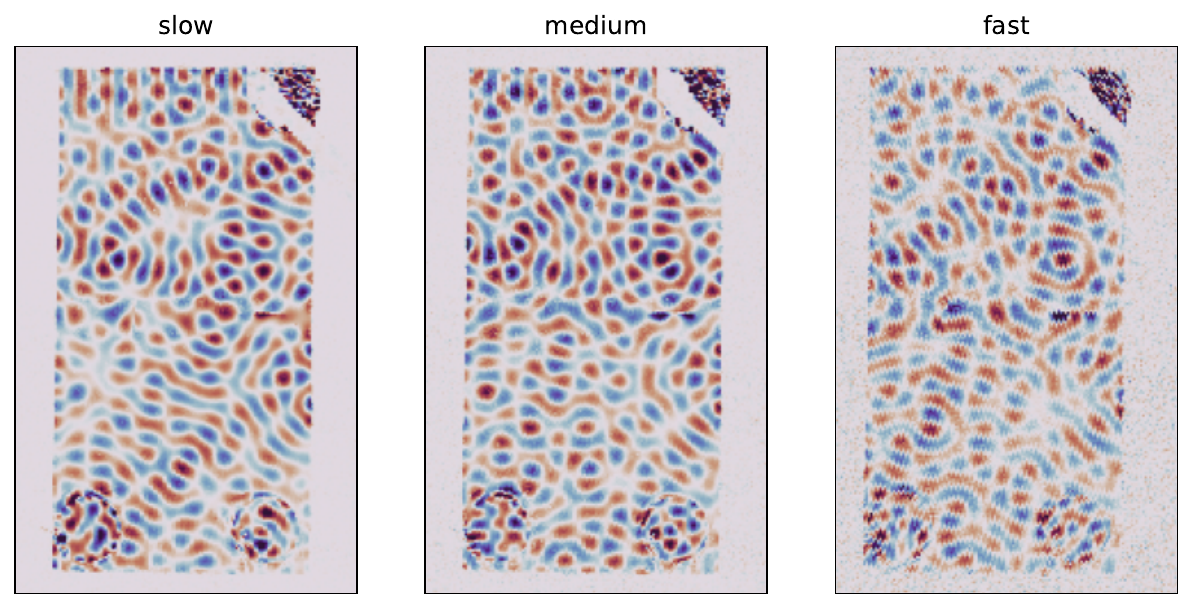}
			\caption{LDV scans at 250 kHz}
	\end{subfigure}
	\caption{The (a) experimental weld sample of 0.25-in.~thick steel with labeled crack regions penetrating roughly 30\% of the weldine, and (b) surface velocities collected by slow, medium, and fast LDV scans. } 
	\label{Fig:ExpSample}
\end{figure}

A single experimental sample was procured from Flawtech with a ground-truth crack defect map (Fig.~\ref{Fig:ExpSample}(a)). This will serve as the validation sample to verify model performance on real-world measurements. 
Two ultrasonic transducers were epoxied onto the bottom of coupon to apply time-harmonic forcing at a steady-state frequency of 250 kHz. 
An LDV measured the surface velocity in a raster pattern; the scans were partitioned and processed to produce an amplitude and phase for the 250 kHz frequency bin in a discretized array, similar to the methodology described in~\cite{Jeon2017}.
We consider three measurement speeds: slow (25 seconds per scan), medium (15 seconds per scan), and fast (5 seconds per scan), as shown in Fig~\ref{Fig:ExpSample}(b). 
The various scan speeds may be used to evaluate the trade-off between measurement time and inverse model performance---this is particularly relevant as a main motivation for steady-state ultrasonic inspection is the significant time reduction as compared to time-domain methods, or more broadly other NDE techniques such as radiography.

\section{Machine Learning Framework} 
\label{sec:ML}

The machine learning methodology addressed two tasks: wavefield inversion and distribution alignment. 
The former is a set of inverse models that are responsible for ingesting wavefield data, either directly from simulation test sets, real-world experiments, or from the output of a conditional generation.
The latter serves to draw out-of-distribution (OOD) experimental samples into the training distribution, i.e., the EM and NL simulations, via conditional generative modeling.

\subsection{Inversion Models}
\label{subsec:Inverse_Models}

The inversion workflow consists of a parallel set of U-Net models to produce estimates of weldline stiffness and to detect defects.
Namely, a spatial inversion U-Net, $\unet_{\inv}$, tasked with ingesting wavefield representations and producing the spatial stiffness map on a 2D domain with respect to the nominal modulus, $\phi(\xEM)=E(\xEM)/E_0$, and a segmentation U-Net, $\unet_{\seg}$, tasked with producing a binary crack detection map defined as
\begin{equation}
	\mathcal{C}=T_\tau[C(\xEM)]= 
	\begin{cases}
		1 & \text{if} \ C(\xEM)\leq\tau\\
		0 & \text{otherwise},
	\end{cases}
\end{equation}
where $\tau$ is one-half the total crack depth if one exists in the sample.

Both models follow the same baseline architecture and data flow. 
A filtering step first extracts the independent wavemodes which, along with the unfiltered wavefield, are passed into the network.
The input wavefield set is denoted $\mathbf{\mathsf{X}} = \left\{ 
\{\wavefield\}\concat
\{\wavefield^{(A,0)}\} \concat
\{\bar\wavefield^{(S,0)} \} 
\right\}$ where $\concat$ is channel-wise concatenation and $\{\wavefield\}$ denotes the scalar fields produced by splitting the complex wavefields into real, imaginary, and absolute components,  $\{\wavefield\} = \left\{
\re(\wavefield),\im(\wavefield),|\wavefield|\right\}$ where again $\wavefield$ denotes $\bm{u}_z(\xEM)$ and $\psi_{\text{EM}}(\xEM)$ interchangeably, as both data sets are used in training.
The independent wavemode component $\wavefield^{(S,0)}$ and $\wavefield^{(A,0)}$ are extracted via wavenumber domain Gaussian filtering as described in section~\ref{subsec: EM_NL_Comparison}.

The baseline architectures of $\unet_{\inv}$ and $\unet_{\seg}$ are both comprised of 5 encoding and decoding blocks, with convolution block attention (CBAM) layers being applied after each convolution block. 
Aside from this, $\unet_{\seg}$ incorporates batch normalization to stabilize training, as well as a multi-axis feature extraction (MAFE) module following CBAM to improve spatial contextualizing in the segmentation task, whereas $\unet_{\inv}$ incorporates Fourier Neural Operator (FNO) modules between skipped connections to capture global dependencies efficiently in the frequency domain, facilitating accurate modeling of complex acoustical fields.
The workflow draws inspiration from the recent results of Ref~\cite{Li2025}, which demonstrated the advantages of FNO, CBAM, and MAFE modules for forward modeling of discontinuous parametric domains.
The details of each module are detailed below. 

\begin{figure}[t!]
	\centering
	\includegraphics[width=\linewidth]{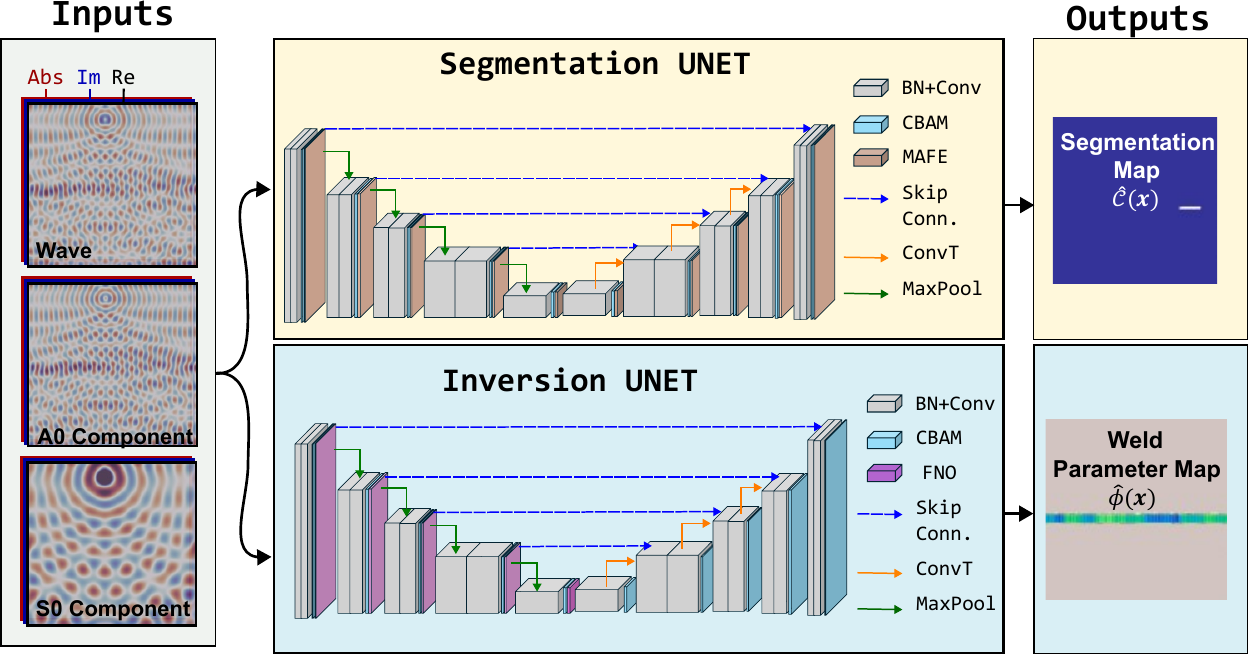}
	\caption{The inverse modeling machine learning framework. Wavefield representations are input as the real, imaginary, and absolute components of the surface displacement, $\wavefield$, and the two filtered Lamb modes. Segmentation and inversion U-Nets operate in parallel to produce $\hat{\mathcal{C}}$ and $\hat\phi$, respectively, using sequence of encoding and decoding blocks detailed for each module with conv and BN denoting convolutional and batch-normalization layers, respectively.
	}
\end{figure}

\subsubsection{Convolutional Block Attention Module (CBAM)}
Given an intermediate feature map tensor, $\feature$, channel attention maps (denoted $ \feature^{\mathsf{C}}$), are computed and subsequently processed along the spatial axis to produce the final attended feature maps, $ \feature^{\mathsf{A}}$. 
Channel attention is first computed by aggregating spatial features via average and max pooling, producing feature vectors encoding summary information on each axis. A shared multi-layer perceptron (MLP), denoted $f_m$, applies nonlinear transformations to the aggregated features, which are then summed and passed through a sigmoid activation to compute attention weights,
\begin{equation}
 \mathsf{W}^{\mathsf{C}} = \mathsf{sigmoid}
 \left(
f_{m}\left(\feature_{\mathsf{MP:S}}\right)  + 
f_{m}\left(\feature_{\mathsf{AP:S}}\right),
\right) 
\end{equation}
with $\feature_{\mathsf{MP:S}}$ and $\feature_{\mathsf{AP:S}}$ denoting spatial max- and average-pooled feature maps.
The map $\feature$ is then scaled channel-wise,
\begin{equation}
	\feature^{\mathsf{C}} = \mathsf{W}^{\mathsf{C}}  \otimes \feature,
\end{equation}
which is then passed to spatial attention module to provide spatial feature importance over $\feature^{\mathsf{C}}$.
It does so by average and max pooling along the channel axis to produce a 2D feature map from $	\feature^{\mathsf{C}}$, which are then passed through a single convolution layer with a $7\times7$ kernel, denoted $f_c$, before applying a final sigmoid activation,
\begin{equation}
	\mathsf{W}^{\mathsf{S}} = \mathsf{sigmoid}
	\left(
	f_{c}\left(
	\feature^{\mathsf{C}}_{\mathsf{MP:C}} \concat
   \feature^{\mathsf{C}}_{\mathsf{AP:C}}  
   \right) 
	\right) 
\end{equation}
with $\feature^{\mathsf{C}}_{\mathsf{MS:C}}$ and $\feature^{\mathsf{C}}_{\mathsf{MP:C}}$ denoting channel-wise max- and average-pooled channel-attended feature maps. 
The attended feature maps are produced by spatially scaling $\feature^{\mathsf{C}}$, 
\begin{equation}
\feature^{\mathsf{A}} = \mathsf{W}^{\mathsf{S}} \otimes	\feature^{\mathsf{C}}
\end{equation}
This sequential channel-then-spatial design allows CBAM to exploit both \emph{what} and \emph{where} to focus on in the feature space, with minimal additional parameters and computational overhead.

\begin{figure}\centering
	\includegraphics[width=.65\linewidth]{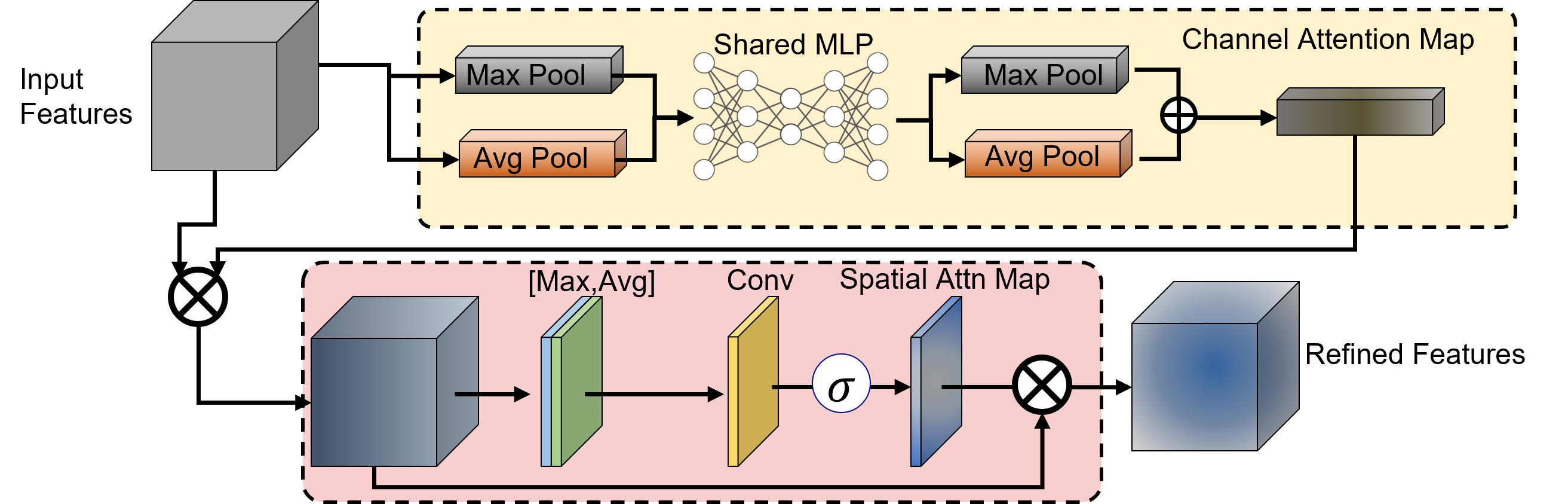}
	\caption{Convolution Block Attention Module (CBAM).}
\end{figure}

\subsubsection{Fourier Neural Operator (FNO)}
FNOs provide a framework for learning mappings between functions by combining integral kernel representations with Fourier transforms~\cite{Li2020}; it has shown efficacy for physics-based processes, such as forward~\cite{Li2025,Wang2024} and inverse modeling of acoustic phenomena~\cite{Yang2021,Zhu2023}.
The FNO data-flow is as follows.
Given a sequence of feature fields $\{\mathbf{\feature}_k\}_{k=0}^K$ on a domain $D \in\domain^{\reduced}$, the update at layer $k$ is defined as,
\begin{equation}
	\begin{aligned}
	\mathbf{\feature}_{k+1}(\xEM)
	&= \sigma\!\left(
O\,\mathbf{\feature}_{k}(\xEM)
	+ (\mathcal{K}_{\theta}\,\mathbf{\feature}_{k})(\xEM)
	\right), \\
	&= 
	\sigma\!\left(
	O\,\mathbf{\feature}_{k}(\xEM)
	+ \int_{D} \eta_{\theta}(\xEM,\xEM^\prime)\,\mathbf{\feature}_{k}(\xEM^\prime)\,d\xEM^\prime
	\right) 
	\end{aligned}
\end{equation}
where $\sigma$ denotes a nonlinear activation function, $O$ is a learned pointwise linear transformation across feature channels, and $\mathcal{K}_{\theta}$ is an integral operator parameterized by a kernel $\eta_{\theta}$, with $\xEM^\prime$ being an integration variable.
Rather than evaluating this convolution directly in the spatial domain, the FNO implements it in the Fourier domain for computational efficiency. 
Defining $R_{\theta}$ as a set of learnable Fourier multipliers, 
the operator can be expressed as
\begin{equation}
	(\mathcal{K}_{\theta}\,\mathbf{\feature}_{k})(\xEM)
	= \mathcal{F}_{\bm{k}}^{-1}\!\Big(
	R_{\theta}(\bm{k})\,
	\mathcal{F}_{\xEM}(\mathbf{\feature}_{k})(\bm{k})
	\Big)(\xEM).
\end{equation}
In practice, the FNO layer combines a local linear transformation in the spatial domain with a global convolution in the Fourier domain, followed by a nonlinear activation. 
This construction enables efficient learning of operators that capture both local and long-range interactions.

\subsubsection{Multi-Axis Feature Extraction}
The Multi-Axis Feature Extraction (MAFE) module is designed to capture both global and local dependencies along spatial axes in order to strengthen contextual representation for segmentation; similar methodologies has proven effective in various image segmentation tasks~\cite{Liao2024,Shao2025}. Herein, we utilize the same MAFE module described Ref~\cite{Li2025}.
Given an input feature map $\feature \in \mathbb{R}^{B \times C \times H \times W}$, where $B$, $C$, $H$, and $W$ denote batch, channel, height, and width, respectively, MAFE applies axis-wise gated MLPs along the horizontal ($f_{m_x}$) and vertical ($f_{m_i}$) directions after normalization, producing a global response, $
\mathsf{g}=f_{m_x}(\feature) + f_{m_y}(\feature)$.
To complement this with local information, the feature map is partitioned into non-overlapping blocks of size $\tfrac{H}{G} \times \tfrac{W}{G}$ ($G=2$ herein), unfolded into patches, and processed by local axis MLPs $f^l_x, f^ly$. This yields the local response
\begin{equation}
	\mathsf{L} = f^l_x(\feature_b) + f^l_y(\feature_b), \quad \feature_b \in \mathbb{R}^{B \cdot G^2 \times C \times \tfrac{H}{G} \times \tfrac{W}{G}}.
\end{equation}
The local responses are then folded back to the original spatial resolution and combined with the global pathway:
 $u = 	\mathsf{g}+ \text{Fold}(	\mathsf{L})$. 
Finally, a projection refines the aggregated output, $\feature_o = W_p\ast u$, 
where $W_p$ is a $1 \times 1$ convolution. 
In this way, MAFE adaptively integrates global context and local structural cues along multiple axes, leading to richer feature representations that improve boundary preservation and semantic consistency in segmentation.

\subsubsection{Inverse Model Training Objectives}
Three competing objectives guide the weld parameter regression model. 
The first is an MSE loss between the true and predicted parameter map on $\xEM$, denoted $\phi(\xEM)$ and scaled by $\mathsf{w}_1$---its purpose is self-evident. 
Second, a gradient loss encourages smoothness in non-welded (flat) regions, which is scaled by $\mathsf{w}_2$.
Finally, a focal loss scaled by $\mathsf{w}_3$  exploits local and spatially compact variations. The composite loss is thus,
\begin{equation}
	\begin{aligned}
		\mathcal{L}_{\inv} =& \mathsf{w}_1\frac{1}{N}\sum_{n=1}^{N}\left\| \phi_n(\xEM)-\hat\phi_n(\xEM) \right\|^2 
		+ \mathsf{w}_2\frac{1}{N}\sum_{n=1}^{N}\left\| \nabla\cdot\phi_n(\xEM)-\nabla\cdot\hat\phi_n(\xEM) \right\|^2
		\\&
		+  
		\mathsf{w}_3\frac{1}{N} \sum_{n=1}^N \left[ \left\| \phi_n(\xEM) - \hat{\phi}_n(\xEM) \right\|^2 
		\left( 1 - \left\| \phi_n(\xEM) - \hat{\phi}_n(\xEM) \right\| \right)^{2}\right].
	\end{aligned}
	\label{EQ:Linv}
\end{equation}
The scaling quantities of Eq~\eqref{EQ:Linv} were selected ad-hoc as $\mathsf{w}_1=1$, $\mathsf{w}_2 = 0.1$, and $\mathsf{w}_3=0.25$ to provide a similar scaling in terms of magnitude. 

The segmentation network is updated by a pair of loss functions aimed at provided a binary segmentation, $\mathcal{C}$, delineating crack regions. 
The first is the usual binary cross-entropy scaled by $\mathsf{w}_4$, with a positive condition weighting factor of 25. The second is a dice loss scaled by $\mathsf{w}_5$. This is a modification of an intersection over union (IOU---a standard loss for segmentation) that more heavily emphasizes true positives (in the current case, a true positive indicates a crack region). This is particularly useful for domains dominated by the null condition, such as detecting small cracks in an otherwise large domain. The combined training objective is written as,
\begin{equation}
	\mathcal{L}_{\text{seg}}  
	=\mathsf{w}_4 \left[ - \frac{1}{N} \sum_{i=1}^N \left( \mathcal{C}_i \log \hat{\mathcal{C}}_i + (1 - \mathcal{C}_i) \log (1 - \hat{\mathcal{C}}_i) \right) \right] 
	+ \mathsf{w}_5 \left[ 1 - \frac{2 \sum_{i=1}^N \hat{\mathcal{C}}_i \, \mathcal{C}_i}{\sum_{i=1}^N \hat{\mathcal{C}}_i + \sum_{i=1}^N \mathcal{C}_i} \right].
	\label{EQ:Lseg}
\end{equation}
The quantities $\mathsf{w}_4$ and $\mathsf{w}_5$ were set constant to 1 and 35, respectively, providing similar magnitude scaling between the two.

\subsection{Super-resolution Conditional Diffusion Model}
\label{subsec:DDPM}

The inversion framework outlines in section~\ref{subsec:Inverse_Models} is exclusively trained on simulated datasets. Ergo, generalizing its performance in a zero-shot fashion to experimental measurements---for which data is too scarce to train with---must be achieved.
Experimental LDV meta-parameters alter the underlying distribution of noise and artifacts in unpredictable ways, making it difficult to conventionally denoise experimental measurements or to synthetically noise training data to represent the diversity of noise distributions that may be present in live deployment. %
Moreover, even the high-fidelity NL solutions are simplified in the sense that they neglect nonlinearity and anisotrpy, and thus miss the full complexity of real-world measurements, meaning that even clean signals require distribution re-alignment when processing experimental data.
Therefore, we propose an alternative to conventional data pre-processing---a generative process that is trained to produce wavefield representations from the training (simulation) distribution based on sparse or corrupted wavefield representations from physical measurements. Specially, we consider a conditioned denoising diffusion probabilistic model (DDPM), detailed herein.

The goal of DDPMs is to learn a generative process that produces realistic samples by reversing a discretized stochastic noise flow. It does so by (i) defining a forward noising process that perturbs data with Gaussian noise according to a fixed schedule, and (ii) training a neural network to approximate the reverse denoising process. Once trained, new samples are generated by starting from Gaussian noise and iteratively applying the learned reverse process. Importantly, DDPM generations can be conditioned on auxiliary input data, such as noisy wavefields, enabling the production of high-fidelity samples consistent with physics-based constraints and indicative of simulation-grade measurements.

Our DDPM workflow follows conventional literature~\cite{Ho2020}.   
The forward process is a Markov chain that progressively adds Gaussian noise to a clean sample form the training distribution, i.e., simulated wavefields, denoted $\wavefieldDDPM_0$. 
The additive noise at each step is determined by a schedule, $\{\beta_t\}_{t=1}^T$, producing the conditional distribution at successive time steps,
\begin{equation}
	q(\wavefieldDDPM_t \mid \wavefieldDDPM_{t-1}) = \mathcal{N}\!\left(\sqrt{1-\beta_t}\,\wavefieldDDPM_{t-1},\, \beta_t \mathbf{I}\right).
\end{equation}
Utilizing the Markovian and Gaussian properties of $\{\beta_t\}_{t=1}^T$, $\wavefieldDDPM_t$ can be queried at any $t$ via,
\begin{equation}
	\wavefieldDDPM_t = \sqrt{\bar{\alpha}_t}\,\wavefieldDDPM_0 + \sqrt{1-\bar{\alpha}_t}\,\boldsymbol{\epsilon}, 
	\qquad \boldsymbol{\epsilon}\sim\mathcal{N}(\mathbf{0},\mathbf{I}),
	\label{EQ:ForwardNoise}
\end{equation}
where $\alpha_t = 1-\beta_t$ and $\bar{\alpha}_t = \prod_{s=1}^t \alpha_s$.

The generative procedure seeks to learn the reverse distribution $p(\wavefieldDDPM_{t-1}\mid \wavefieldDDPM_t)$, i.e., the conditional probability of a slightly less noisy sample given a current noisy sample,
\begin{equation}
	p_\theta(\wavefieldDDPM_{t-1} \mid \wavefieldDDPM_t) = 
	\mathcal{N}\!\big(\wavefieldDDPM_{t-1};\, \boldsymbol{\mu}_\theta(\wavefieldDDPM_t, t),\, \sigma_t^2 \mathbf{I}\big),
\end{equation}
where $\boldsymbol{\mu}_\theta$ is predicted by a neural network and $\sigma_t^2$ follows the fixed variance schedule.  
Under Gaussian assumptions, the mean can be expressed as
\begin{equation}
	\boldsymbol{\mu}_\theta(\wavefieldDDPM_t, t) = 
	\frac{1}{\sqrt{\alpha_t}}\left(\wavefieldDDPM_t - 
	\frac{1-\alpha_t}{\sqrt{1-\bar{\alpha}_t}}\,\boldsymbol{\epsilon}_\theta(\wavefieldDDPM_t,t)\right),
\end{equation}
where $\boldsymbol{\epsilon}_\theta$ denotes the network’s prediction of $\boldsymbol{\epsilon}$.
Training is accomplished by minimizing a variational bound on the data likelihood. In practice, this reduces to a denoising score-matching style objective:
\begin{equation}
	\mathcal{L}_{\text{simple}} = 
	\mathbb{E}_{t,\,\wavefieldDDPM_0,\,\boldsymbol{\epsilon}}
	\Big[ \big\| \boldsymbol{\epsilon} - \boldsymbol{\epsilon}_\theta\!\left( 
	\sqrt{\bar{\alpha}_t}\wavefieldDDPM_0 + \sqrt{1-\bar{\alpha}_t}\boldsymbol{\epsilon},\, t 
	\right)\big\|^2 \Big].
	\label{EQ:Lsimple}
\end{equation}
This objective encourages the network to accurately predict the injected noise, thereby implicitly learning the data distribution. We utilize the attention U-Net available from the \texttt{deeplay}\footnote{\href{https://github.com/DeepTrackAI/deeplay}{\texttt{https://github.com/DeepTrackAI/deeplay}}} repository with 5 blocks and sinusoidal positional encoding. 

\subsubsection{Conditioned DDPM}
\begin{figure}[t!]\centering
	\begin{subfigure}{\linewidth} 
		\includegraphics[width=\linewidth]{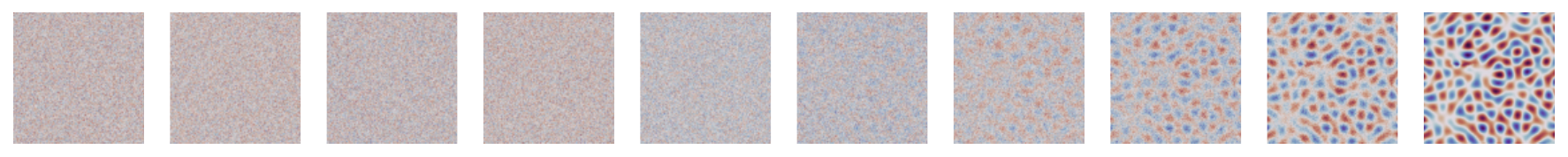}
		\caption{Reverse diffusion process moving from left $(\wavefield_T)$ to right $(\wavefield_0)$ with snap shots of evenly distributed $p_\theta(\wavefieldDDPM_{t-1}|\wavefieldDDPM_t)$ for $t\in[0,1000]$.}
	\end{subfigure}
	\begin{subfigure}{\linewidth}  
		\includegraphics[width=\linewidth]{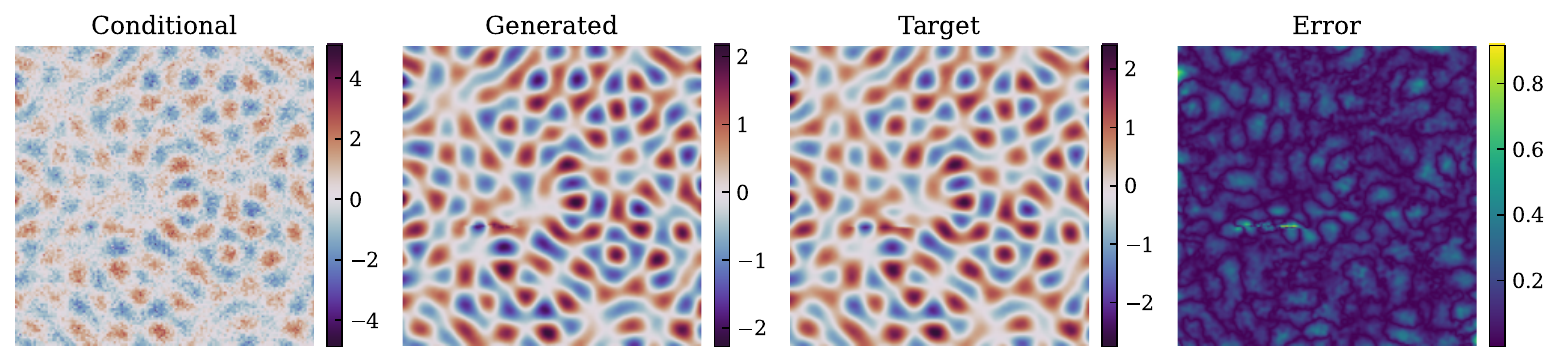}  
		\caption{Conditional generation of a target wavefield depicting $\tilde\wavefieldDDPM$, $p_{\theta}(\wavefieldDDPM_0| {\wavefieldDDPM^{\rm exp}_{\rm synth}})$, target wavefield $\wavefieldDDPM_0$, and error $|p_{\theta}(\wavefieldDDPM_0| {\wavefieldDDPM^{\rm exp}_{\rm synth}})-\wavefieldDDPM_0|$ from left-to-right. }
	\end{subfigure}
	\caption{Example of the DDPM reverse diffusion process showing (a) the reverse process from $\wavefieldDDPM_{T}$ to $\wavefieldDDPM_0$, and (b) a conditional generation of a target simulation wavefield.}
	\label{FIG:DDPM}
\end{figure}	
In many applications, including wavefield reconstruction, corrupted and noisy observations of $\wavefield$ are available (i.e., an experimental scan), which we denote $\wavefield^{\rm exp}$. The forward process may therefore be conditioned on such inputs, guiding the reverse process to generate outputs consistent with observed data. This conditioning greatly enhances the fidelity of the reconstructions, enabling DDPM to produce samples that are not only realistic but also consistent with simulation-grade measurements. 
Herein, we utilize a simple conditioning approach and concatenate auxiliary to the model input along the channel dimension at every time step.
The DDPM model is trained using conditional data that is taken to be a synthetically corrupted (noised) version of simulation wavefields, 
\begin{equation}
	\wavefield^{\rm exp}_{\rm synth} = \left[\wavefieldDDPM + 
	\mathcal{N}(\mathbf{0},\varepsilon\sigma_{\wavefieldDDPM}\mathbf{I})
	 + \varepsilon \sigma_{\wavefieldDDPM}S(\xEM) \right]\odot \textbf{M},
	\label{EQ:AddedNoise}
\end{equation}
where $\sigma_{\wavefieldDDPM}$ is the noise level nominally set to be 50 percent the variance of $\wavefieldDDPM$, $\varepsilon\sim\mathcal{N}(1,0.5)$ ensures a diversity of noise levels are utilized for conditional wavefields, 
$S(\xEM)$ is a speckle-noise pattern, and $\mathbf{M}$ is a masking function that removes 25 percent of pixels at random and is applied 50 percent of the time in training---this prevents the DDPM from becoming too-reliant on conditional data and neglecting $\wavefieldDDPM_{t-1}$ and the $t$ during training.

Fig~\ref{FIG:DDPM} depicts the reverse diffusion process showing the evolution from $\wavefieldDDPM_T$ to $\wavefieldDDPM_0$ with various snap-shots of $p_\theta(\wavefieldDDPM_{t-1}|\wavefieldDDPM_t)$ (Fig~\ref{FIG:DDPM}(a)), as well as a conditional generation of a target wavefield (Fig~\ref{FIG:DDPM}(b)). Notably, in Fig~\ref{FIG:DDPM}(b), the conditionally generated wavefield   is shown to be qualitatively representative of the target simulation, with little error between the two. Thus, the conditional generation procedure provides a method for generated in-distribution representations of occluded or corrupted wavefields with high fidelity and accuracy. 
In section~\ref{subsubsec:DDPM_Guide}, we expand adapt this procedure to enable OOD distribution alignment of experimental measurements.

\begin{figure}[t!]\centering
	\includegraphics[width=\linewidth]{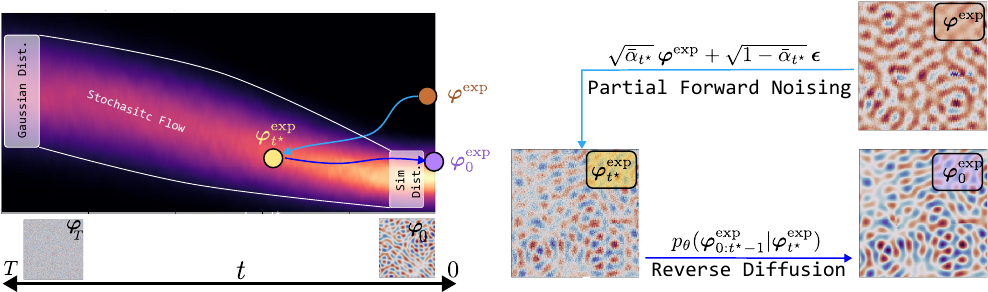}  
		\caption{The guided reverse diffusion process for experimental measurements. The colormap represents a simplification of the stochastic flow guiding the reverse diffusion process from a Gaussian $\wavefieldDDPM_T$ to the training (simulation) distribution, $\wavefieldDDPM_0$. OOD experimental data $\wavefieldEX$ is forward diffused onto the flow at time $t$, $\wavefieldEX_{\tstar}$, which then follows the trajectory by to time $t=0$.
		}
		\label{FIG:DDPM_Ex}
\end{figure}	

\subsubsection{Guided DDPM for Experimental Wavefields}
\label{subsubsec:DDPM_Guide}
We utilize reverse diffusion to align OOD experimental measurements to simulation distributions as follows.
Defining $\wavefieldEX$ as an LDV scan collected at either a slow, medium, or fast rate, we apply the forward DDPM noising flow per Eq~\eqref{EQ:ForwardNoise} to time $t^\star<T$.
This provides a partially noised representation of $\wavefieldEX$, denoted $\wavefieldEX_\tstar$---it is qualitatively similar to the distribution of partially noised training signals at a similar diffusion time step $\tstar$, e.g., $\wavefield_{\tstar}$.
We utilize $\wavefieldEX_\tstar$ as the starting point for an abbreviated reverse diffusion process beginning at $t=\tstar$ and ending at $t=0$ to produce $p_\theta(\wavefieldEX_{0:\tstar-1}|\wavefieldEX_\tstar)$, where $\wavefieldEX_0$ is a generated sample that is based on $\wavefieldEX$, but now within the training distribution, e.g., a NL simulation counterpart to  $\wavefieldEX$.

The conditional data used to guide the reverse diffusion is taken as the measurement itself, $\wavefieldEX$.
A pictorial representation of this process is given by Fig~\ref{FIG:DDPM_Ex}, showing a mock 1D rendition of the DDPM reverse stochastic flow with $\wavefieldEX$ initially laying off the simulation distribution. Injecting noise to $\wavefieldEX_{\tstar}$ places the sample onto the stochastic flow learning in training, from which condition reverse diffusion guides it to $\wavefieldEX_0$---a simulation-grade rendition of $\wavefieldEX$.

\subsection{Enrichment Training Scheme}
\label{subsec:Training}
The training schemes of Eqs~\eqref{EQ:Linv}, \eqref{EQ:Lseg}, and~\eqref{EQ:Lsimple}  utilize a mixture of inexpensive EM and high-fidelity NL solutions in training.
The large distributions of EM solutions serve to train models over a diversity of patterns, whereas the NL solutions are used to tune the models to more-realistic wave solutions. 
In this sense, the EM solutions serve enrich the training distribution
The following epoch-dependent scaling function controls the evolving focus between the two data distributions, 
\begin{equation}
	\mathcal{L}^{\text{joint}}_{m}(\bm\vartheta) = \varpi_{\EM}(e) \mathcal{L}^{\EM}_{m}(\bm\vartheta; \mathbf{\mathsf{X}}_{\EM}) + \varpi_{\NL}(e) \mathcal{L}^{\NL}_{m}(\bm\vartheta; \mathbf{\mathsf{X}}_{\NL}).
	\label{EQ:LossSched}
\end{equation}
Here,  $\varpi_{\EM}(e)=1-e$ and $\varpi_{\NL}(e)=1+e$ are the scaling constants of the EM and NL losses, respectively, denoted as $\mathcal{L}^{\EM}_{m}$ and $\mathcal{L}^{\NL}_{m}$ for corresponding datasets $\mathbf{\mathsf{X}}_{\EM}$ and $\mathbf{\mathsf{X}}_{\NL}$, and $e$ is the relative epoch number ranging between 0 and 1. 
The result of Eq~\eqref{EQ:LossSched} is a training scheme that emphasizes EM solutions at the beginning of training, and gradually shifts focus to NL solutions as the models are tuned. 
Although the loss weights start off equal, the over-abundance of EM data still leads to a strong emphasis on EM losses unless $e\ll1$.
Moreover, as focus gradually transitions to NL data, the models retain the need to capture the broad distributions of EM solutions while emphasizing the limited NL distributions---this is preferably to traditional transfer learning which risks over-fitting to the narrow NL distributions and forgetting the broader diversity of the EM dataset.

For the inverse and segmentation models, an exponential linear decay schedule is applied with constants of 0.98 and 0.95, respectively. 
Each model is trained for 100 epochs with the exception of the diffusion models, which are trained for 150 epochs.  
In all training runs, and 80-20 split of training and testing data is implemented, with the data-splits  held consistent for each model.


\section{Results} 
\label{sec:results}

We utilize numerical and physical experiments to evaluate the enrichment training scheme, the model performance on hold-out NL simulations, and the adaptability to real-world data.
We first demonstrate the EM solutions (\ref{subsec: EM_Model}) provide meaningful data enrichment for learning the inverse of high-fidelity NL models (\ref{subsec: NL_Model}).
This is followed by qualitative and quantitative performance metrics for each of the problem classes described in section~\ref{subsec:domains}. 
Lastly, the performance of the DDPM-driven distribution alignment described by section~\ref{subsec:DDPM} and downstream inverse task is shown for the experimental data described in section~\ref{subsec:ExMeas}.

\subsection{Computational Results}

\begin{figure}[t!]
	\includegraphics[width=\linewidth]{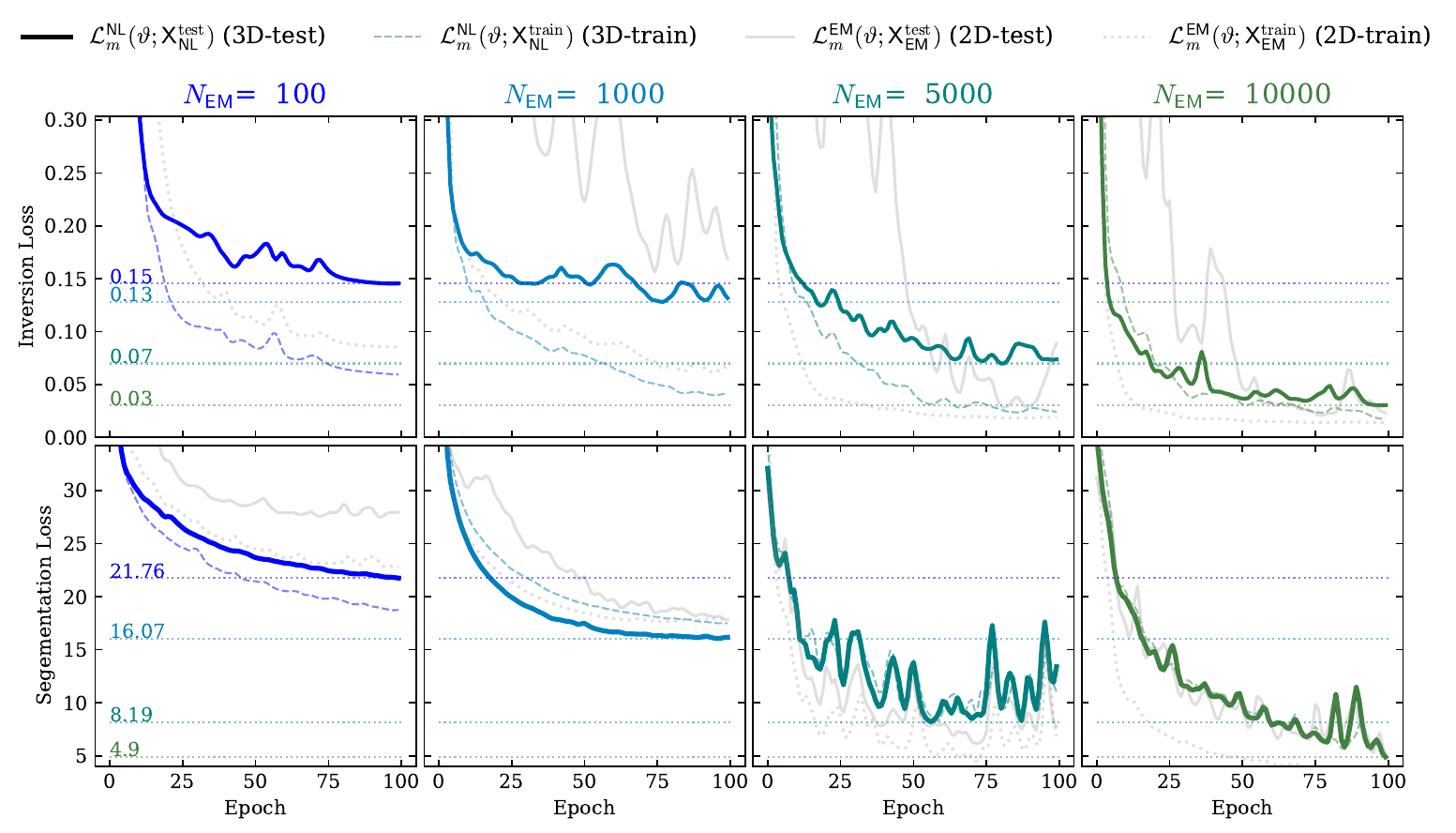}
	\caption{The test-loss for high-fidelity NL solutions over 100 epochs of training as the number of enrichment solutions (e.g., EM solutions) available in training is increased from 100 to 10,000 samples. 
	The top row depicts the inversion loss (Eq~\eqref{EQ:Linv}) while the bottom row depicts segmentation loss (Eq~\eqref{EQ:Lseg}).
	}
	\label{FIG:Loss}
\end{figure}


\subsubsection{Effective Medium Solution Enrichment}
We may determine the value of EM solutions for learning the NL inverse by varying the number of available EM training simulations and evaluating the corresponding NL test loss.
The results are summarized by Fig~\ref{FIG:Loss} for four quantities of EM enrichment solutions: $N_{\EM}=$ 100, 1,000, 5,000, and 10,000. 
Solid colored curves represent the NL test loss, $\mathcal{L}^{\NL}_{m}(\bm\vartheta; \mathbf{\mathsf{X}}_{\NL}^{\text{test}})$---this evaluation data is held consistent between the trials.
These are accompanied by colored dashed curves representing NL training loss, $\mathcal{L}^{\NL}_{m}(\bm\vartheta; \mathbf{\mathsf{X}}_{\NL}^{\text{train}})$, as well as solid and dashed gray curves representing the training and test losses of EM solutions $\mathcal{L}^{\EM}_{m}(\bm\vartheta; \mathbf{\mathsf{X}}_{\EM}^{\text{train}})$ and $\mathcal{L}^{\EM}_{m}(\bm\vartheta; \mathbf{\mathsf{X}}_{\EM}^{\text{test}})$. 
The horizontal dashed lines indicated the lowest achieved test loss during training for each enrichment dataset size, emphasizing the quantitative magnitude in loss reduction as $N_{\EM}$ grows.

From Fig~\ref{FIG:Loss}, it is clear that incorporating more EM enrichment solutions improves performance on high-fidelity NL test simulations.
There are two key takeaways from this result. 
First, that the datasets generated by Eqs~\eqref{EQ:EffectiveMediumPDE} and~\eqref{EQ:EffectiveMediumSol} are qualitatively representative of the high fidelity solutions given by Eq~\eqref{EQ:force_balance}, e.g., that the EM model captures the dominant scattering features on a significantly reduced computational cost.
Secondly, it demonstrates that the enrichment training scheme of Eq.~\eqref{EQ:LossSched} successfully transfers knowledge from the distribution of reduced-order EM solutions to the comparatively limited set of high-fidelity 3D simulations.

\subsubsection{Qualitative Model Evaluation}

\begin{figure}
	{\tiny }	\includegraphics[width=\linewidth]{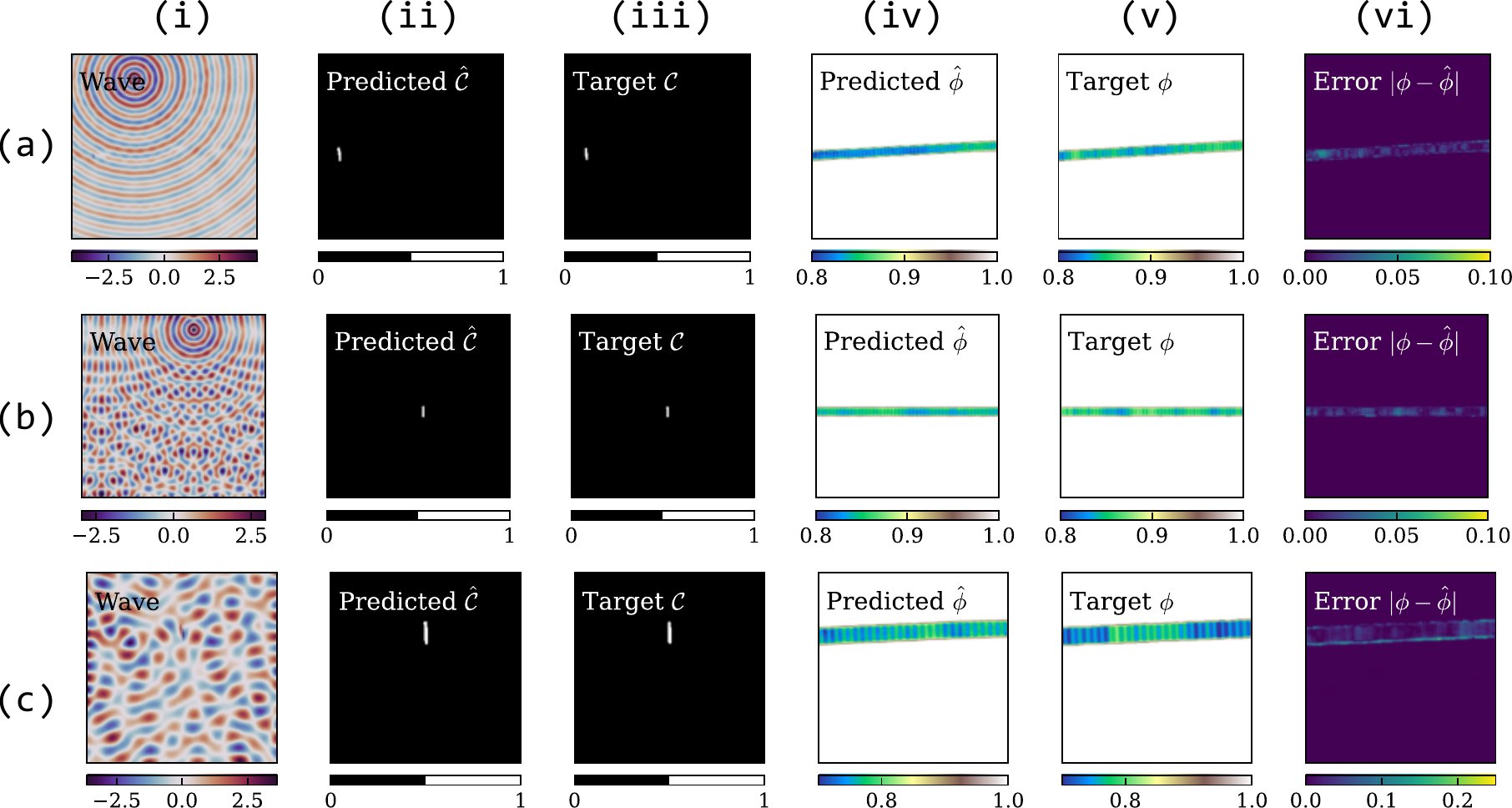} 
	\caption{qualitative examples of model inference for (a) scattering, (b) periodic, and (c) free-free boundary conditions depicting the (i) real value of the wavefield $\bm{u}_z(x,y,z=h)$, (ii) the predicted crack location, (iii) the true crack locations, (iv) the predicted weld stiffness variations, (v) the true weld stiffness variations, and (vi) the disagreement between (v) and (vi).}
	\label{FIG:qualitative_eval}
\end{figure}

The qualitative performance of the inverse models on a randomly selected test simulation from each problem class (e.g., scattering, periodic, or free-free) is given by Fig~\ref{FIG:qualitative_eval}.
The surface response $\bm{u}_z(\xEM)$ is given in column (i), representing  one of  the nine inverse model inputs. 
Columns (ii)-(vi) compare the predicted versus ground truth inverse solutions both for crack segmentation, $\mathcal{C}(\xEM)$, and  spatially-resolved domain stiffness variation $\phi(\xEM)$.
Agreement is achieved between predicted and true crack masks, both in terms of spatial location and relative size. 
Qualitative agreement is recovered for weld parameter inference as well, though this is difficult to confirm visually. However, the relatively low error indicated by column (vi) implies that the weldlines are well-characterized for each example.

\subsubsection{Quantitative Model Evaluation}
Quantitative performance analysis is conduced using 200 hold-out high fidelity simulations of each problem class.
For weld characterization, we consider average error in the normalized stiffness reduction over weld regions as a performance metric.
The error analysis neglects bulk regions (i.e., $\domain_b$)---incorporating this would skew the results favorably in the models favor as predicting the nominal stiffness is a simple regression task.
To quantify crack identification performance, we consider the classical true(false) positive(negative) paradigm to evaluate effectiveness on the binary detection task. 
We note that this is a relatively well-understood image processing task for which many classical methods exist---performance should be very good for binary crack detection on clean wavefields. 
Crack length characterization is a more challenging task, which we evaluate this by regressing the predicted versus actual crack length.

\begin{figure}[t!]
	\includegraphics[width=\linewidth]{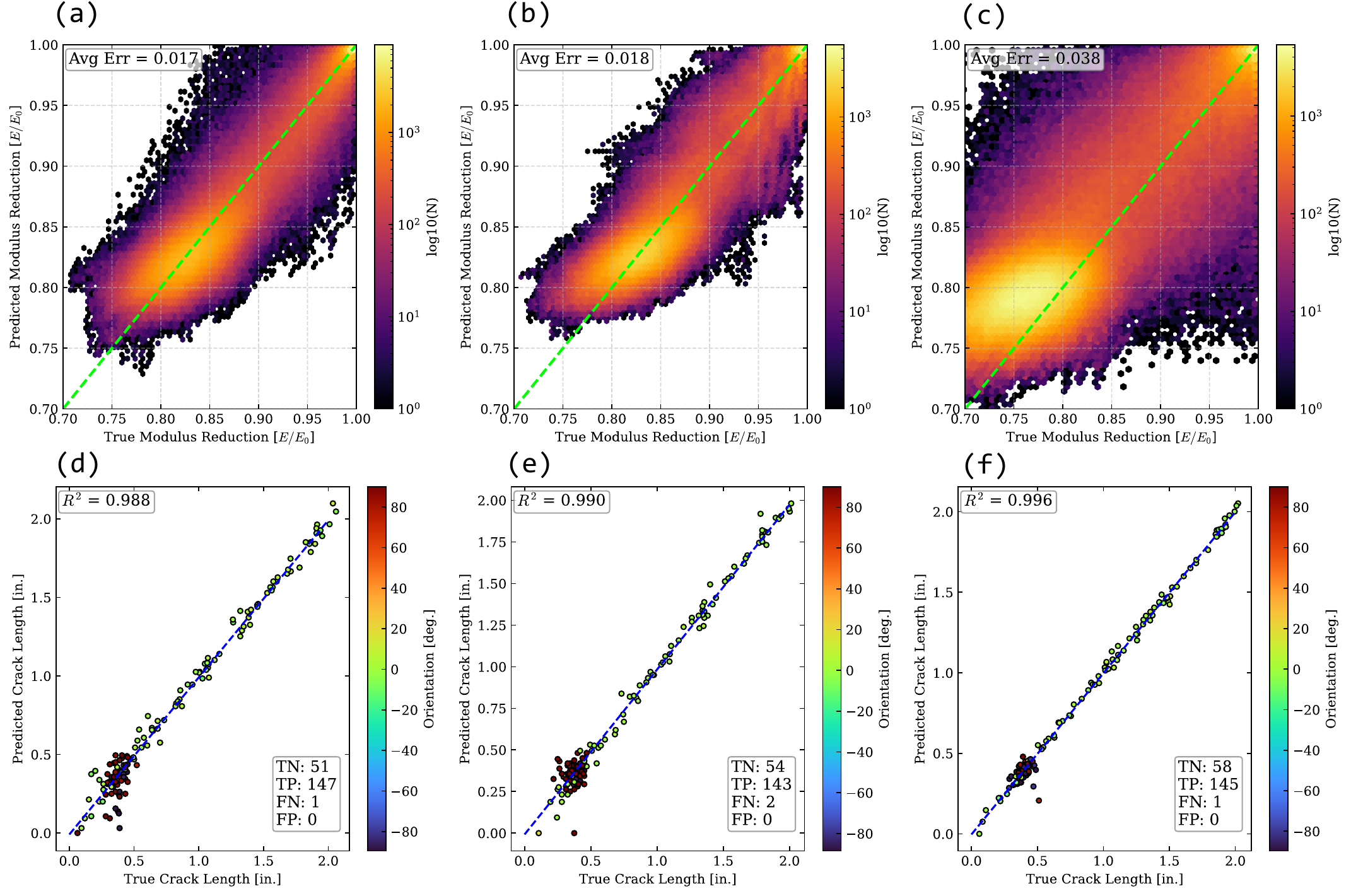}
	\caption{The statistical summary of weld parameter inference for (a) scattering, (b) periodic, and (c) free-free problem classes described by Eqs~\eqref{EQ:BC_scatter}-\eqref{EQ:BC_coupon} and Fig~\ref{Fig:Probs}. (e-f) The corresponding statistical summary of crack predictions are given directly below.}
	\label{FIG:quantitative_eval}
\end{figure}

The quantitative evaluation results are given in Fig~\ref{FIG:quantitative_eval}. 
Subplots \ref{FIG:quantitative_eval}(a)-\ref{FIG:quantitative_eval}(c) depict the true versus predicted stiffness reduction in weldline regions for the scattering, periodic, and free-free problem classes, respectively. 
A reasonable correlation is achieved for each, with the predictions following reasonably the axis diagonal (green dashed lines).
The free-free problem classes seemingly displays the broadest deviations from the target, with an average error roughly double that of the other problem types.
This is a result of the increased complexity introduced by the reflections from each exterior boundary, making the interpretation of localized wave patterns more nuanced compared to the scattering or periodic problems which each incorporate PML layers. Moreover, the coupon problem class was parameterized with a larger nominal reduction as compared to the other two.

Subplots~\ref{FIG:quantitative_eval}(d)-\ref{FIG:quantitative_eval}(f) depict the crack characterization results for the corresponding problem domain of subplots (a)-(c).
The breakdown of binary detection measures are shown in the bottom-right of each subplot.
Near perfect detection is achieved for each problem domain, with only one or two false negatives reported and no false positives. 
The more challenging task---crack characterization---is represented by the scatter plots of true versus predicted crack length. 
A near perfect $R^2$ score is recovered over each domain, indicating very good predictive capacity for characterizing crack length. 
There is a noticeable clustering of lengths around the 0.25 to 0.5 inch range---these represent cracks that occur normal to the weldline (which typically span the weld width).
The crack orientations are given by the colorbar with little correlation to predictive performance indicated. 

The summary findings of Fig~\ref{FIG:quantitative_eval} are as follows. 
The inverse model successfully predicts the weld stiffness reduction within reasonable accuracy, though error is non-negligible.
In contrast, the modeling framework is far more effective in crack identification and characterization, with little error observed. 
As mentioned, the former task poses a greater challenge than the latter, and thus the discrepancy in performance is expected among these tasks. 
Nevertheless, this study demonstrates successful performance on test data generated by high-fidelity elastodynamic simulations.

\subsection{Experimental Results}

\begin{figure}[t!]
	\includegraphics[width=\linewidth]{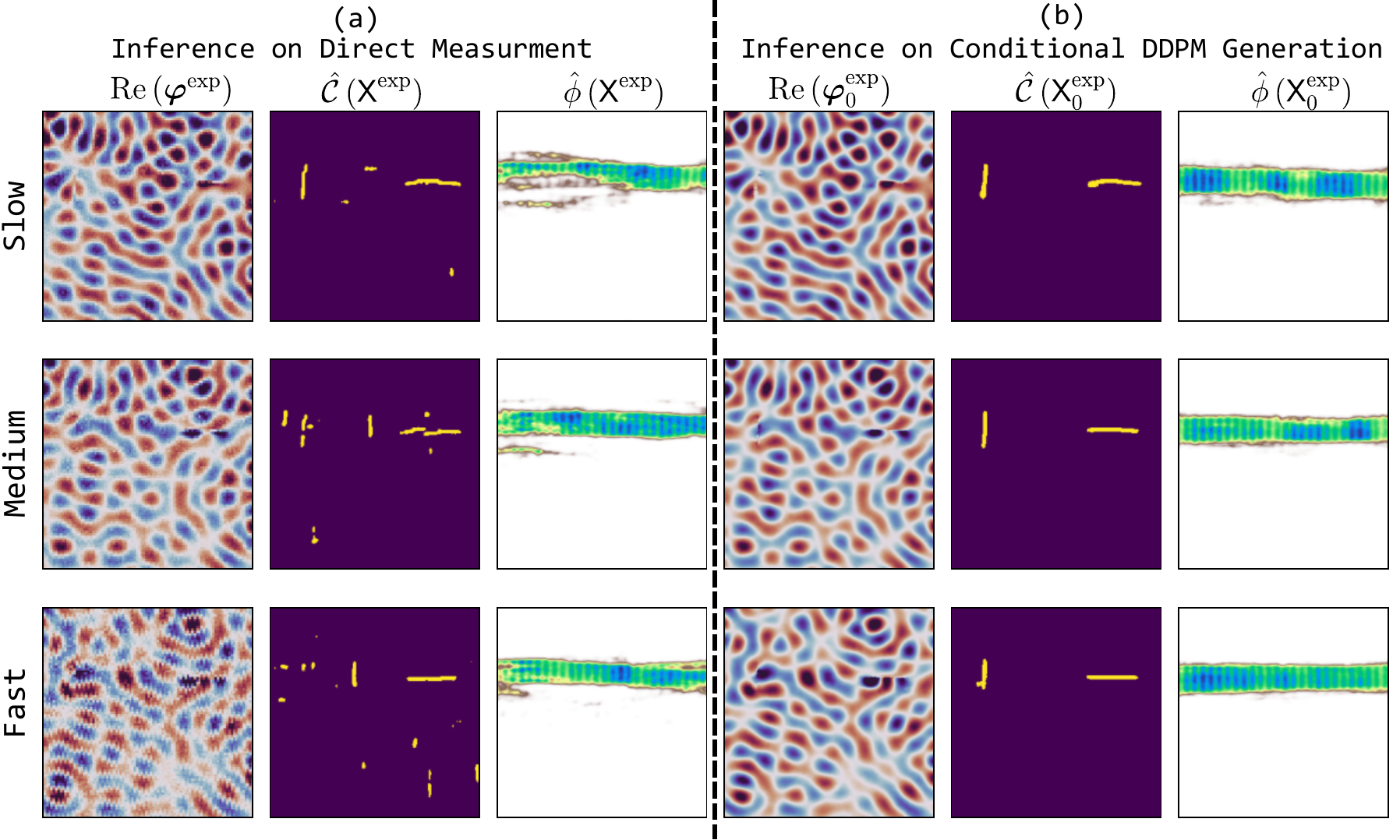}  
	\caption{The inversion results on experimental measurements with the rows corresponding to scanned at slow, medium, and fast scan speeds from top-to-bottom.
	Results for (a) direct measurements are given in the left panel, depicting the real-part of the experimental input wave $\wavefield^{\text{exp}}$, as well as crack and weld parameter inference on the input wavefield set, $\hat{\mathcal{C}}\left(\mathsf{X}^{\text{exp}}\right)$ and  $\hat{\phi}\left(\mathsf{X}^{\text{exp}}\right)$, respectively.
	The same results are given for (b) conditional generations of based on the measured inputs, with subscript $\square_0$ indicating that the field represents to the reverse diffusion generation at the final time-step, $t=0$. 
	}
	\label{FIG:ExpResults}
\end{figure}

To demonstrate the model's capacity for predicting properties of rapid on-the-fly surface wavefield measurements---a critical need to make the methodology practical---we consider the model performance on the experimental measurements described in section~\ref{subsec:ExMeas}.
Herein, the notation $\mathsf{X}^{\text{exp}}$ denotes the same set of nine model input wavefields described in section~\ref{subsec:Inverse_Models}, but for the experimentally observed scans depicted by Fig~\ref{Fig:ExpSample}(b)  cropped to a 4$\times$4 inch region centered near the weld.
In addition to inference on the direct measurements, conditionally generated samples for each scan are considered as distribution-aligned variant of the direct measurements, denoted $\mathsf{X}^{\text{exp}}$.

The inference results on direct measurements are given by Fig~\ref{FIG:ExpResults}(a), showing the real-component of $\wavefieldEX$ for each scan speed as well as the model's crack and weld stiffness predictions on $\mathsf{X}^{\text{exp}}$.
The models struggle to achieve either weld characterization or crack identification on the direct measurements.
Although the large crack of the experimental sample is identified for all scan speeds, it is accompanied by many false positives. 
As the scan rate increases, the smaller crack is lost, and the predicted crack maps seem to lose all meaning with respect to the ground truth depicted in Fig~\ref{Fig:ExpSample}. 
Moreover, the parameter maps produced by $\mathsf{U}_{\rm inv}$ are difficult to interpret, with discontinuities and a non-physically noisy boundaries shown. 
Hence, as expected, the model fails to generalize to OOD experimental measurements directly.

Fig~\ref{FIG:ExpResults}(b) depicts the inverse same results applied to conditionally generated fields, $\wavefieldEX_0$.
 Now, in contrast, exceptional performance is recovered for the crack detection problem, with the predicted mask $\hat{\mathcal{C}}\left(\mathsf{X}^{\text{exp}}_0\right)$ indicating cracks in the same two regions flagged by the calibrated ground-truth mask of Fig~\ref{Fig:ExpSample}.
 Moreover, the inference for the weld properties are now consistent with the expected result: nominal bulk properties in the unwelded domain, and a noticeable (15-20 percent) stiffness reduction with in the weldline itself. 
 Hence, the proposed framework distribution alignment of OOD experimental measurements to the known training distribution (based on 3D elastodynamic modeling) is shown effective for applying the inverse model to real-world measurements.

\section{Conclusions and Future Work}
\label{sec:conclusions}

In this work, we have presented a machine learning framework for the automated inspection of weld stiffness and the detection of cracks.
A high-fidelity elastodynamic simulation framework was first established to generate realistic ultrasonic wavefield responses for materials containing heterogeneous and potentially cracked weldlines. To complement this, an effective medium (EM) model based on Lamb wave theory was developed to efficiently produce a large and diverse training dataset. The inversion modules—specifically, a material inversion U-Net and a crack detection U-Net—were trained on both the reduced-order and high-fidelity datasets in a controlled fashion, thereby enabling knowledge transfer from the efficient EM models to the more computationally expensive simulations. Additionally, a conditional diffusion probabilistic model (DDPM) was trained to generate simulation-grade, in-distribution measurements from corrupted and coarsely sampled real-world data.

We demonstrated that the EM solutions captured the dominant features of the full-fidelity 3D simulations, both qualitatively—by comparing wavefields and spectra—and quantitatively—by showing that including a larger number of EM solutions in training improved model performance on held-out 3D simulations. The efficacy of our modeling approach was further validated across multiple weld inspection scenarios, including scattering problems with full PML boundaries, pipe-problems with PML-periodic mixed conditions, and coupon problems with free-free boundaries. Finally, we showed that the combination of the inversion models with the generative DDPM enabled effective inference on measured weld specimens. By generating simulation-grade representations of experimental scans across different scan speeds and subsequently passing them through the inversion models, we achieved robust inference results that outperformed direct application of the models to out-of-distribution experimental scans.

This framework has the potential to significantly improve the reliability and efficiency of weld inspection by reducing the dependence on large volumes of costly high-fidelity simulations and by bridging the gap between simulation and experimental domains. The integration of generative modeling with inversion networks enables more robust inference under challenging measurement conditions, suggesting promising applicability for industrial-scale nondestructive evaluation. However, the current approach remains limited by the simplifications inherent in the EM models and by the restricted diversity of experimental datasets used for validation. Additionally, the framework’s performance in highly complex geometries or under extreme noise conditions has yet to be systematically assessed. Future work will focus on expanding the scope of experimental validation, incorporating more realistic material heterogeneity and defect morphologies, and further developing physics-informed generative models to enhance robustness and interpretability. In parallel, extending the inversion framework to handle multimodal inspection data (e.g., combining ultrasonic, radiographic, and thermographic methods) presents a promising direction for achieving more comprehensive weld integrity assessment.

\section*{Acknowledgments}
Funding was provided by the Laboratory Directed Research and Development Office through project \#20240779PRD1. 
\\

\bibliographystyle{elsarticle-num}
\bibliography{Refs}

\begin{thebibliography}{10}
\expandafter\ifx\csname url\endcsname\relax
  \def\url#1{\texttt{#1}}\fi
\expandafter\ifx\csname urlprefix\endcsname\relax\def\urlprefix{URL }\fi
\expandafter\ifx\csname href\endcsname\relax
  \def\href#1#2{#2} \def\path#1{#1}\fi

\bibitem{Sun2023}
H.~Sun, P.~Ramuhalli, R.~E. Jacob, Machine learning for ultrasonic
  nondestructive examination of welding defects: A systematic review,
  Ultrasonics 127 (2023) 106854.
\newblock \href {https://doi.org/10.1016/j.ultras.2022.106854}
  {\path{doi:10.1016/j.ultras.2022.106854}}.

\bibitem{Shafeek2004}
H.~Shafeek, E.~Gadelmawla, A.~Abdel-Shafy, I.~Elewa, Automatic inspection of
  gas pipeline welding defects using an expert vision system, NDT \& E
  International 37~(4) (2004) 301--307.
\newblock \href {https://doi.org/10.1016/j.ndteint.2003.10.004}
  {\path{doi:10.1016/j.ndteint.2003.10.004}}.

\bibitem{Kasban2011}
H.~Kasban, O.~Zahran, H.~Arafa, M.~El-Kordy, S.~Elaraby, F.~Abd El-Samie,
  Welding defect detection from radiography images with a cepstral approach,
  NDT \&; E International 44~(2) (2011) 226--231.
\newblock \href {https://doi.org/10.1016/j.ndteint.2010.10.005}
  {\path{doi:10.1016/j.ndteint.2010.10.005}}.

\bibitem{Nadzri2018}
N.~A. Nadzri, M.~Ishak, M.~M. Saari, A.~M. Halil, Development of eddy current
  testing system for welding inspection, in: Proceedings of the 9th IEEE
  Control and System Graduate Research Colloquium (ICSGRC), 2018, pp. 94--98.

\bibitem{Saini1998}
D.~Saini, S.~Floyd, An investigation of gas metal arc welding sound signature
  for on-line quality control, Welding Journal 77~(5) (1998) 172s.

\bibitem{Grad2004}
L.~Grad, J.~Grum, I.~Polajnar, J.~M. Slabe, Feasibility study of acoustic
  signals for on-line monitoring in short circuit gas metal arc welding,
  International Journal of Machine Tools and Manufacture 44~(5) (2004)
  555--561.
\newblock \href {https://doi.org/10.1016/j.ijmachtools.2003.10.016}
  {\path{doi:10.1016/j.ijmachtools.2003.10.016}}.

\bibitem{Zhang2019AE}
L.~Zhang, A.~C. Basantes-Defaz, D.~Ozevin, E.~Indacochea, Real-time monitoring
  of welding process using air-coupled ultrasonics and acoustic emission,
  International Journal of Advanced Manufacturing Technology 101~(5-8) (2019)
  1623--1634.
\newblock \href {https://doi.org/10.1007/s00170-018-3042-2}
  {\path{doi:10.1007/s00170-018-3042-2}}.

\bibitem{Mohandas2024}
R.~Mohandas, P.~Mongan, M.~Hayes, Ultrasonic weld quality inspection involving
  strength prediction and defect detection in data-constrained training
  environments, Sensors 24~(20) (2024) 6553.
\newblock \href {https://doi.org/10.3390/s24206553}
  {\path{doi:10.3390/s24206553}}.

\bibitem{Alonso2022}
J.~Alonso, S.~Pav{\'o}n, J.~Vidal, M.~Delgado, Advanced comparison of phased
  array and x-rays in the inspection of metallic welding, Materials 15~(20)
  (2022) 7108.
\newblock \href {https://doi.org/10.3390/ma15207108}
  {\path{doi:10.3390/ma15207108}}.

\bibitem{Chu2016}
H.~H. Chu, Z.~Y. Wang, A vision-based system for post-welding quality
  measurement and defect detection, International Journal of Advanced
  Manufacturing Technology 86~(9-12) (2016) 3007--3014.
\newblock \href {https://doi.org/10.1007/s00170-015-8334-1}
  {\path{doi:10.1007/s00170-015-8334-1}}.

\bibitem{Zolfaghari2018}
A.~Zolfaghari, A.~Zolfaghari, F.~Kolahan, Reliability and sensitivity of
  magnetic particle nondestructive testing in detecting the surface cracks of
  welded components, Nondestructive Testing and Evaluation 33~(3) (2018)
  290--300.
\newblock \href {https://doi.org/10.1080/10589759.2018.1428322}
  {\path{doi:10.1080/10589759.2018.1428322}}.

\bibitem{Dorafshan2018}
S.~Dorafshan, M.~Maguire, W.~Collins, Infrared thermography for weld
  inspection: feasibility and application, Infrastructures 3~(4) (2018) 45.
\newblock \href {https://doi.org/10.3390/infrastructures3040045}
  {\path{doi:10.3390/infrastructures3040045}}.

\bibitem{Li2011}
T.~Li, D.~P. Almond, D.~A.~S. Rees, Crack imaging by scanning pulsed laser spot
  thermography, NDT \& E International 44~(2) (2011) 216--225.
\newblock \href {https://doi.org/10.1016/j.ndteint.2010.08.006}
  {\path{doi:10.1016/j.ndteint.2010.08.006}}.

\bibitem{Roca2007}
A.~S. Roca, H.~C. Fals, J.~B. Fern{\'a}ndez, E.~J. Mac{\'i}as, F.~S. Ad{\'a}n,
  New stability index for short circuit transfer mode in gmaw process using
  acoustic emission signals, Science and Technology of Welding and Joining
  12~(5) (2007) 460--466.
\newblock \href {https://doi.org/10.1179/174329307X213882}
  {\path{doi:10.1179/174329307X213882}}.

\bibitem{Gaja2017}
H.~Gaja, F.~Liou, Defects monitoring of laser metal deposition using acoustic
  emission sensor, International Journal of Advanced Manufacturing Technology
  90~(1-4) (2017) 561--574.
\newblock \href {https://doi.org/10.1007/s00170-016-9366-x}
  {\path{doi:10.1007/s00170-016-9366-x}}.

\bibitem{Bowler2022}
A.~L. Bowler, M.~P. Pound, N.~J. Watson, A review of ultrasonic sensing and
  machine learning methods to monitor industrial processes, Ultrasonics 124
  (2022) 106776.
\newblock \href {https://doi.org/10.1016/j.ultras.2022.106776}
  {\path{doi:10.1016/j.ultras.2022.106776}}.

\bibitem{Moles2005}
M.~Moles, N.~Dub{\'e}, S.~Labb{\'e}, E.~Ginzel, Review of ultrasonic phased
  arrays for pressure vessel and pipeline weld inspections, Journal of Pressure
  Vessel Technology 127~(3) (2005) 351--356.
\newblock \href {https://doi.org/10.1115/1.1991881}
  {\path{doi:10.1115/1.1991881}}.

\bibitem{Lopez2019}
A.~B. Lopez, J.~Santos, J.~P. Sousa, T.~G. Santos, L.~Quintino, Phased array
  ultrasonic inspection of metal additive manufacturing parts, Journal of
  Nondestructive Evaluation 38~(3) (2019) 60.
\newblock \href {https://doi.org/10.1007/s10921-019-0600-y}
  {\path{doi:10.1007/s10921-019-0600-y}}.

\bibitem{Moreno2013}
E.~Moreno~Hern{\'a}ndez, R.~Otero, B.~Arregi~Landa, N.~Galarza~Urigoitia,
  B.~Rubio~Garcia, Use of lamb waves high modes in weld testing, e-Journal of
  Nondestructive Testing 18, article Vol. 18 (1), available at
  \url{https://www.ndt.net/?id=13781} (2013).

\bibitem{Kundu2004}
T.~Kundu, Ultrasonic Nondestructive Evaluation: Engineering and Biological
  Material Characterization, CRC Press, 2004.

\bibitem{Mirapeix2007}
J.~Mirapeix, P.~Garc{\'\i}a-Allende, A.~Cobo, O.~Conde, J.~M.
  L{\'o}pez-Higuera, Real-time arc-welding defect detection and classification
  with principal component analysis and artificial neural networks, NDT \& E
  International 40~(4) (2007) 315--323.
\newblock \href {https://doi.org/10.1016/j.ndteint.2006.12.001}
  {\path{doi:10.1016/j.ndteint.2006.12.001}}.

\bibitem{You2014}
D.~You, X.~Gao, S.~Katayama, Wpd-pca-based laser welding process monitoring and
  defects diagnosis by using fnn and svm, IEEE Transactions on Industrial
  Electronics 62~(1) (2014) 628--636.
\newblock \href {https://doi.org/10.1109/TIE.2014.2319216}
  {\path{doi:10.1109/TIE.2014.2319216}}.

\bibitem{Khumaidi2017}
A.~Khumaidi, E.~M. Yuniarno, M.~H. Purnomo, Welding defect classification based
  on convolution neural network (cnn) and gaussian kernel, in: 2017
  International Seminar on Intelligent Technology and Its Applications
  (ISITIA), 2017, pp. 261--265.
\newblock \href {https://doi.org/10.1109/ISITIA.2017.8124091}
  {\path{doi:10.1109/ISITIA.2017.8124091}}.

\bibitem{Zhang2019Weld}
Z.~Zhang, G.~Wen, S.~Chen, Weld image deep learning-based on-line defects
  detection using convolutional neural networks for al alloy in robotic arc
  welding, Journal of Manufacturing Processes 45 (2019) 208--216.
\newblock \href {https://doi.org/10.1016/j.jmapro.2019.06.023}
  {\path{doi:10.1016/j.jmapro.2019.06.023}}.

\bibitem{Munir2018Ultras}
N.~Munir, H.-J. Kim, J.~Park, S.-J. Song, S.-S. Kang, Convolutional neural
  network for ultrasonic weldment flaw classification in noisy conditions,
  Ultrasonics 94 (2019) 74--81.
\newblock \href {https://doi.org/10.1016/j.ultras.2018.12.001}
  {\path{doi:10.1016/j.ultras.2018.12.001}}.

\bibitem{Silva2020}
L.~C. Silva, E.~F.~S. Filho, M.~C.~S. Albuquerque, I.~C. Silva, C.~T.~T.
  Farias, Segmented analysis of time-of-flight diffraction ultrasound for flaw
  detection in welded steel plates using extreme learning machines, Ultrasonics
  102 (2020) 106057.
\newblock \href {https://doi.org/10.1016/j.ultras.2019.106057}
  {\path{doi:10.1016/j.ultras.2019.106057}}.

\bibitem{Virkkunen2021}
I.~Virkkunen, T.~Koskinen, O.~Jessen-Juhler, J.~Rinta-aho, Augmented ultrasonic
  data for machine learning, Journal of Nondestructive Evaluation 40~(4) (2021)
  1--11.
\newblock \href {https://doi.org/10.1007/s10921-020-00739-5}
  {\path{doi:10.1007/s10921-020-00739-5}}.

\bibitem{Madhvacharyula2022}
A.~S. Madhvacharyula, V.~S.~P. Araveeti, S.~Gorthi, S.~Chitral, N.~Venkaiah,
  V.~K. Degala, In situ detection of welding defects: A review, Welding in the
  World 66 (2022) 611--628.
\newblock \href {https://doi.org/10.1007/s40194-021-01229-6}
  {\path{doi:10.1007/s40194-021-01229-6}}.

\bibitem{RocaBarcelo2017}
F.~Roca~Barceló, P.~Jaén~del Hierro, F.~Ribes~Llario, J.~Real~Herráiz,
  Development of an ultrasonic weld inspection system based on image processing
  and neural networks, Nondestructive Testing and Evaluation 33~(2) (2017)
  229--236.
\newblock \href {https://doi.org/10.1080/10589759.2017.1376056}
  {\path{doi:10.1080/10589759.2017.1376056}}.

\bibitem{Yuan2024}
Z.~Yuan, X.~Gao, K.~Yang, J.~Peng, L.~Luo, Performance enhancement of
  ultrasonic weld defect detection network based on generative data, Journal of
  Nondestructive Evaluation 43~(4) (2024).
\newblock \href {https://doi.org/10.1007/s10921-024-01119-z}
  {\path{doi:10.1007/s10921-024-01119-z}}.

\bibitem{Tripicchio2020}
P.~Tripicchio, G.~Camacho-Gonz{\'a}lez, S.~D'Avella, Welding defect detection:
  coping with artifacts in the production line, International Journal of
  Advanced Manufacturing Technology 111 (2020) 1659--1669.
\newblock \href {https://doi.org/10.1007/s00170-020-06172-2}
  {\path{doi:10.1007/s00170-020-06172-2}}.

\bibitem{BiasuzBlock2024}
S.~Biasuz~Block, R.~Dutra~da Silva, A.~Eugnio~Lazzaretti, R.~Minetto,
  Lohi-weld: A novel industrial dataset for weld defect detection and
  classification, a deep learning study, and future perspectives, IEEE Access
  12 (2024) 77442--77453.
\newblock \href {https://doi.org/10.1109/access.2024.3407019}
  {\path{doi:10.1109/access.2024.3407019}}.

\bibitem{Handoko2023}
D.~Handoko, Weld defect detection and classification based on deep learning
  method: A review, Jurnal Ilmu Komputer dan Informasi 16~(2) (2023) 77--87.

\bibitem{Koskinen2018}
T.~Koskinen, I.~Virkkunen, S.~Papula, T.~Sarikka, J.~Haapalainen, Producing a
  pod curve with emulated signal response data, Insight 60~(1) (2018) 42--48.

\bibitem{Zhang2025}
L.~Zhang, H.~Pan, B.~Jia, L.~Li, M.~Pan, L.~Chen, Lightweight dcgan and
  mobilenet based model for detecting x-ray welding defects under unbalanced
  samples, Scientific Reports 15 (2025) Article 6221.
\newblock \href {https://doi.org/10.1038/s41598-025-89558-0}
  {\path{doi:10.1038/s41598-025-89558-0}}.

\bibitem{Ho2020}
J.~Ho, A.~Jain, P.~Abbeel,
  \href{https://proceedings.neurips.cc/paper_files/paper/2020/file/4c5bcfec8584af0d967f1ab10179ca4b-Paper.pdf}{Denoising
  diffusion probabilistic models}, in: H.~Larochelle, M.~Ranzato, R.~Hadsell,
  M.~Balcan, H.~Lin (Eds.), Advances in Neural Information Processing Systems,
  Vol.~33, Curran Associates, Inc., 2020, pp. 6840--6851.
\newline\urlprefix\url{https://proceedings.neurips.cc/paper_files/paper/2020/file/4c5bcfec8584af0d967f1ab10179ca4b-Paper.pdf}

\bibitem{Yang2024}
P.~Yang, R.~Zhang, Z.~Li, X.~Wang, Diffusion model based denoising for
  ultrasonic nde signals under industrial noise, Ultrasonics 132 (2024) 107004.
\newblock \href {https://doi.org/10.1016/j.ultras.2023.107004}
  {\path{doi:10.1016/j.ultras.2023.107004}}.

\bibitem{Cantero2022}
S.~Cantero-Chinchilla, P.~D. Wilcox, A.~J. Croxford, Deep learning in automated
  ultrasonic nde--developments, axioms and opportunities, Ndt \& E
  International 131 (2022) 102703.
\newblock \href {https://doi.org/10.1016/j.ndteint.2022.102703}
  {\path{doi:10.1016/j.ndteint.2022.102703}}.

\bibitem{Lhemery2000}
A.~Lh{\'e}mery, P.~Calmon, I.~Lec{\oe}ur-Ta{\i}bi, R.~Raillon, L.~Paradis,
  Modeling tools for ultrasonic inspection of welds, NDT\&E International
  33~(7) (2000) 499--513.
\newblock \href {https://doi.org/10.1016/s0963-8695(00)00021-9}
  {\path{doi:10.1016/s0963-8695(00)00021-9}}.

\bibitem{Hu2019}
M.~Hu, J.~Li, Exploring bias in gan-based data augmentation for small samples
  (May 2019).
\newblock \href {http://arxiv.org/abs/1905.08495} {\path{arXiv:1905.08495}},
  \href {https://doi.org/10.48550/ARXIV.1905.08495}
  {\path{doi:10.48550/ARXIV.1905.08495}}.

\bibitem{Giurgiutiu2014}
V.~Giurgiutiu, Guided Waves, Elsevier, 2014, pp. 293--355.
\newblock \href {https://doi.org/10.1016/b978-0-12-418691-0.00006-x}
  {\path{doi:10.1016/b978-0-12-418691-0.00006-x}}.

\bibitem{Jeon2017}
J.~Y. Jeon, S.~Gang, G.~Park, E.~Flynn, T.~Kang, S.~Woo~Han, Damage detection
  on composite structures with standing wave excitation and wavenumber
  analysis, Advanced Composite Materials 26~(sup1) (2017) 53--65.
\newblock \href {https://doi.org/10.1080/09243046.2017.1313577}
  {\path{doi:10.1080/09243046.2017.1313577}}.

\bibitem{Li2025}
C.~Li, H.~Zhao, Y.~Hao, A feature enhanced autoencoder integrated with fourier
  neural operator for intelligent elastic wavefield modeling, IEEE Transactions
  on Geoscience and Remote Sensing 63 (2025) 1--16.
\newblock \href {https://doi.org/10.1109/tgrs.2025.3542082}
  {\path{doi:10.1109/tgrs.2025.3542082}}.

\bibitem{Li2020}
Z.~Li, N.~Kovachki, K.~Azizzadenesheli, B.~Liu, K.~Bhattacharya, A.~Stuart,
  A.~Anandkumar, Fourier neural operator for parametric partial differential
  equations (Oct. 2020).
\newblock \href {http://arxiv.org/abs/2010.08895} {\path{arXiv:2010.08895}},
  \href {https://doi.org/10.48550/ARXIV.2010.08895}
  {\path{doi:10.48550/ARXIV.2010.08895}}.

\bibitem{Wang2024}
Y.~Wang, H.~Zhang, C.~Lai, X.~Hu, Transfer learning fourier neural operator for
  solving parametric frequency-domain wave equations, IEEE Transactions on
  Geoscience and Remote Sensing 62 (2024) 1--11.
\newblock \href {https://doi.org/10.1109/tgrs.2024.3440199}
  {\path{doi:10.1109/tgrs.2024.3440199}}.

\bibitem{Yang2021}
Y.~Yang, A.~F. Gao, J.~C. Castellanos, Z.~E. Ross, K.~Azizzadenesheli, R.~W.
  Clayton, Seismic wave propagation and inversion with neural operators, The
  Seismic Record 1~(3) (2021) 126--134.
\newblock \href {https://doi.org/10.1785/0320210026}
  {\path{doi:10.1785/0320210026}}.

\bibitem{Zhu2023}
M.~Zhu, S.~Feng, Y.~Lin, L.~Lu, Fourier-deeponet: Fourier-enhanced deep
  operator networks for full waveform inversion with improved accuracy,
  generalizability, and robustness, Computer Methods in Applied Mechanics and
  Engineering 416 (2023) 116300.
\newblock \href {https://doi.org/10.1016/j.cma.2023.116300}
  {\path{doi:10.1016/j.cma.2023.116300}}.

\bibitem{Liao2024}
J.~Liao, H.~Wang, H.~Gu, Y.~Cai, Liver tumor segmentation method combining
  multi-axis attention and conditional generative adversarial networks, PLOS
  ONE 19~(12) (2024) e0312105.
\newblock \href {https://doi.org/10.1371/journal.pone.0312105}
  {\path{doi:10.1371/journal.pone.0312105}}.

\bibitem{Shao2025}
H.~Shao, Q.~Zeng, Q.~Hou, J.~Yang, Mcanet: Medical image segmentation with
  multi-scale cross-axis attention, Machine Intelligence Research 22~(3) (2025)
  437--451.
\newblock \href {https://doi.org/10.1007/s11633-025-1552-6}
  {\path{doi:10.1007/s11633-025-1552-6}}.

\end{thebibliography}

\appendix

\clearpage
\newpage

\section{Weldline Parameterization Functions}  
\label{APX: Weld_functions}

The bead function is selected to provide realistic impedance profiles around the weld-line. It's upon the path $\Gamma_c$ and radius function $f_r(d_\perp)$, where $d_\perp$ is the perpendicular distance from the weld path,
\begin{equation}
	d_\perp = \left( \begin{bmatrix} x \\ y \end{bmatrix} - \gamma(s^*) \right) \cdot \mathbf{n}(s^*)
\end{equation}
where $\gamma(s^*)$ is the nearest point of a coordinate to the weld path, and $\textbf{n}(s^*)$ a normal vector.
A semi-circle protrusion about the weld path is then modeled via,
\begin{equation}
	f(d_\perp) = 
	\begin{cases}
		\sqrt{1 - \left(\frac{d_\perp}{R}\right)^2}, & |d_\perp| \leq R - r_f \\
		\sqrt{1 - \left(\frac{d_\perp}{R}\right)^2} \cdot \frac{1 + \cos\left( \pi \cdot \frac{|d_\perp| - (R - r_f)}{r_f} \right)}{2}, & R - r_f < |d_\perp| \leq R \\
		0, & |d_\perp| > R
	\end{cases}
	\label{EQ:Arc2d}
\end{equation}
where $R$ is the weld radius and $r_f-\alpha R$ a small fillet radius to ensure smoothness in the domain.
Moreover, real welds have more appreciable impedance mismatch along weldline boundaries than those modeled by~\eqref{EQ:Arc2d}, and we therefore consider an additional boundary wavespeed reduction term as,
\begin{equation}
	f_b(s) = \begin{cases}
		1, & d_i\leq r_0-w\\
		\frac{1}{2}\cos((\pi d_i-(r_0-w))/w) & r_0-w<d_i\leq r_0\\
		0, & d_i>r_0
	\end{cases}
\end{equation}
The bead function ${B}(s)$ is subsequently applied as
\begin{equation}
	B_k(s) = A_k\cdot\left(
	1-\left(\frac{s-s_k}{\sigma_b}
	\right)^2\right)^p, \ \ |s-s_k|\leq\sigma_b,
\end{equation}
where $A_k = a(1+\varepsilon_k)$ is a pseudo-random bead height and $s_k$ the pseudo-random position of the $k$-th bead along $\Gamma_w$ given by  $s_{k+1}=s_k+s_0(1+\mathcal{U}(-\beta,\beta))$.
To account for variations in weld thickness, which is common in real welding applications and likely to bias our model against robust crack detection if not accounted for, we consider a smoothed randomized Gaussian field of magnitude $\varepsilon_V$
\begin{equation}
	V(s) = \mathcal{G}\ast \mathcal{N}(0,1)\varepsilon_V. 
\end{equation}

\clearpage
\newpage

\section{Finite Element Implementation}
\label{APX:FEMS}
\subsection{Elastodynamic Equations}
The elastodynamic model of Eq~\eqref{EQ:force_balance} was solved via a Galerkin-FE approximation.  
Let $V_h\subset V:=\{\bm{v}\in[H^1(\domain)]^3| \bm{v}=0 \ \text{on} \ \dirichlet \}$ be the vector-valued quadratic ($P_2$) Lagrange space with homogeneous Dirichlet conditions on $\dirichlet$. We seek a trial function $\bm{u}_h\in V_h$ such that for all $\bm{v}_h\in V_h$,
\begin{equation}
	\int_\Omega \frac{1}{J}\,\tilde{\boldsymbol{\sigma}}(\bm{u}_h):\tilde{\boldsymbol{\varepsilon}}(\bm{v}_h)\,{\rm d}  \domain
	\;-\;\omega^2 \int_\Omega \rho\,\bm{u}_h\cdot\bm{v}_h\,{\rm d}  \domain
	\;=\;\int_\Omega \mathbf f\cdot\bm{v}_h\,{\rm d}  \domain
	\;+\;\int_{\Gamma_t}\mathbf t\cdot\bm{v}_h\,{\rm d}\Gamma 
\end{equation}
Choosing a basis $\{\vectorBasis_i\}_{i=1}^n$ to approximate $\bm{u}_h$ on the function space leads to the formation of elemental and global mass and stiffness matrices with a corresponding force vector.
Let $\mathcal{E}_e$ be an element with local scalar $P_2$ shape functions $\{N_a\}_{a=1}^{n_e}$ and spatial dimension $d\in\{2,3\}$.
Defining the vector shape matrix $\mathbf N(x)=\mathrm{blkdiag}(N_1 I_d,\ldots,N_{n_e} I_d)$, strain–displacement matrix $\tilde{\mathbf B}(x)$ built from $\tilde{\nabla}$, and the isotropic elasticity matrix $\mathbf D(x)$ formed from  the Lam\'e parameters interpolated onto the mesh,
the assembly of elemental matrices are given as,
\begin{align}
	\textbf{K} &=  \mathop{\mathsf{A}}\limits_{e=1}^{n_\mathrm{el}} \left(
	\int_{\mathcal{E}_e} \frac{1}{J}\;\tilde{\mathbf B}^\top\,\mathbf D\,\tilde{\mathbf B}\; {\rm d}  \domain\right), \label{eq:Ke}\\
	\textbf{M} &=  \mathop{\mathsf{A}}\limits_{e=1}^{n_\mathrm{el}} \left(  \int_{\mathcal{E}_e} \rho\;\mathbf N^\top \mathbf N\; {\rm d}  \domain\right), \label{eq:Me}\\
	\textbf{F} &=  \mathop{\mathsf{A}}\limits_{e=1}^{n_\mathrm{el}} \left( \int_{\mathcal{E}_e} \mathbf N^\top \mathbf f\; {\rm d}  \domain \;+\; \int_{\partial \mathcal{E}_e\cap\Gamma_t} \mathbf N^\top \mathbf t\; d\Gamma\right). \label{eq:Fe}
\end{align} 
with $\mathsf{A}$ being the assembly operator. $\mathbf{t}=\mathbf{0}$ for all problems considered herein.
After imposing Dirichlet boundary conditions on $\dirichlet$, the discrete solution $\bm{u}$ is obtained from frequency-domain matrix-vector equation,
\begin{equation}
	\left(\textbf{K}-\omega^2 \textbf{M}\right)\,\bm{u} = \mathbf F,
	\label{EQ:EvalProbl_Vec}
\end{equation}
and the finite element field follows as $\bm{u}_h = \sum_{j=1}^n U_j \vectorBasis_j$.

\subsection{Helmholtz Equation}
Let $W_h \subset W = \{v \in H^1(\Omega^\reduced) : v= 0 \text{ on } \dirichlet^\reduced \}$ 
be the scalar quadratic ($P_2$) Lagrange space with homogeneous conditions on $\dirichlet^\reduced$. 
We seek $\psi_h^{(m,n)} \in W_h$ such that for all $v_h \in W_h$
\begin{equation}\begin{multlined} 
		\int_{\Omega^\reduced} \frac{\kappa^{(m,n)}(\bar{\bm{x}})}{\xi(\bar{\bm{x}})} 
		\nabla_{\!\bar{\bm{x}}}\psi_h^{(m,n)} \cdot \nabla_{\!\bar{\bm{x}}} v_h \, {\rm d}  \domain^\reduced
		\;+\;\int_{\Omega^\reduced} \frac{\omega^2}{\xi(\bar{\bm{x}})} \, \psi_h^{(m,n)} v_h \, {\rm d}  \domain^\reduced
		\\
		= \int_{\Omega^\reduced} f^{(m,n)} v_h \, {\rm d}  \domain^\reduced
		+ \int_{\Gamma_N^\reduced} g^{(m,n)} v_h \, ds. 
	\end{multlined}
\end{equation}
where  $\kappa^{(m,n)}(\xEM) =\left( \Phi^{(m,n)}(\xEM)\cdot v_p^{(m,n)}\cdot C(\xEM) \right)^2$ and $g^{(m,n)}=0$ for all problems considered herein.
Approximating $\psi_h^{(m,n)}$ on the basis $\{\scalarBasis_i\}_{i=1}^{N}$ for $W_h$ such that 
$\psi_h^{(m,n)} = \sum_{j=1}^N U_j^{(m,n)} \scalarBasis_j$, we may derive a similar frequency-domain matrix-vector equation for each mode $(m,n)$.
Let $\mathcal{E}_e$ be an element with local quadratic scalar shape functions $\{N_a\}_{a=1}^{n_e}$ 
and spatial dimension $d=2$. 
Define the element shape function and gradient matrices on $W_h$ as $\bar{\mathbf{N}}$ and $\bar{\mathbf{B}}$, the global matrices and load vector are given by
\begin{align}
	\mathbf K^{(m,n)} &= 
	\mathop{\mathsf A}\limits_{e=1}^{n_{\mathrm{el}}}
	\!\left(\int_{\mathcal{E}_e} \frac{\kappa^{(m,n)}(\bar{\bm{x}})}{\xi(\bar{\bm{x}})} \; \bar{\mathbf{B}}^{\!\top}\bar{\mathbf{B}}\; {\rm d}  \domain^\reduced\right),
	\\[6pt]
	\mathbf M^{(m,n)} &= 
	\mathop{\mathsf A}\limits_{e=1}^{n_{\mathrm{el}}}
	\!\left(\int_{\mathcal{E}_e} \frac{1}{\xi(\bar{\bm{x}})} \; \bar{\mathbf{N}}^{\!\top}\bar{\mathbf{N}} \; {\rm d}  \domain^\reduced\right),
	\\[6pt]
	\mathbf F^{(m,n)} &= 
	\mathop{\mathsf A}\limits_{e=1}^{n_{\mathrm{el}}}
	\!\left(\int_{\mathcal{E}_e} \bar{\mathbf{N}}^{\!\top} f^{(m,n)} \, {\rm d}  \domain^\reduced
	\;+\; \int_{\partial \mathcal{E}_e \cap \Gamma_N^\reduced} \bar{\mathbf{N}}^{\!\top} g^{(m,n)} \, ds\right). 
\end{align}
After imposing Dirichlet boundaries on $\dirichlet^\reduced$, the discrete solution for each mode $\mathbf U^{(m,n)}$ is then obtained from its corresponding matrix-vector equation,

\begin{equation}
	\left(\textbf{K}^{(m,n)}-\omega^2 \textbf{M}^{(m,n)}\right)\,\mathbf{U}^{(m,n)} = \textbf{F}^{(m,n)}.
	\label{EQ:EvalProbl_Sca}
\end{equation}

\subsection{Solution}
Solutions to Eqs~\eqref{EQ:EvalProbl_Vec} and~\eqref{EQ:EvalProbl_Sca}  were obtained using the PETSc library. 
In this setting, the Krylov solver was employed with direct LU factorization preconditioning---this was carried out using the parallel sparse direct solver MUMPS with customized pivot thresholding and minimum pivot size parameters. 
With this setup, the system matrix $\mathbf{A} =(\mathbf{K}-\omega^2\mathbf{M})$ was factorized once, after which the solution vector 
$\mathbf{U}$ is obtained directly from the right-hand side 
$\mathbf{F}$ through the factorized operator.

\clearpage
\newpage

\section{Comparison of Conditioned DDPM to DnCNN}

\label{subsec:dncnn}

The Denoising Convolutional Neural Network (DnCNN) is a deep feed-forward architecture designed for image denoising by learning a residual mapping. Instead of directly predicting the clean image \( \hat{x} \), DnCNN is trained to estimate the noise \( \hat{v} \) present in the noisy observation \( y \), where \( y = x + v \). The network learns to minimize the residual loss \( \mathcal{L} = \| \hat{v} - (y - x) \|_2^2 \), thereby recovering the denoised image as \( \hat{x} = y - \hat{v} \). DnCNN typically uses a series of convolutional layers with batch normalization and ReLU activations, enabling it to model complex noise patterns efficiently while maintaining computational simplicity.

\begin{figure}[t!]\centering
	\includegraphics[width=.95\linewidth]{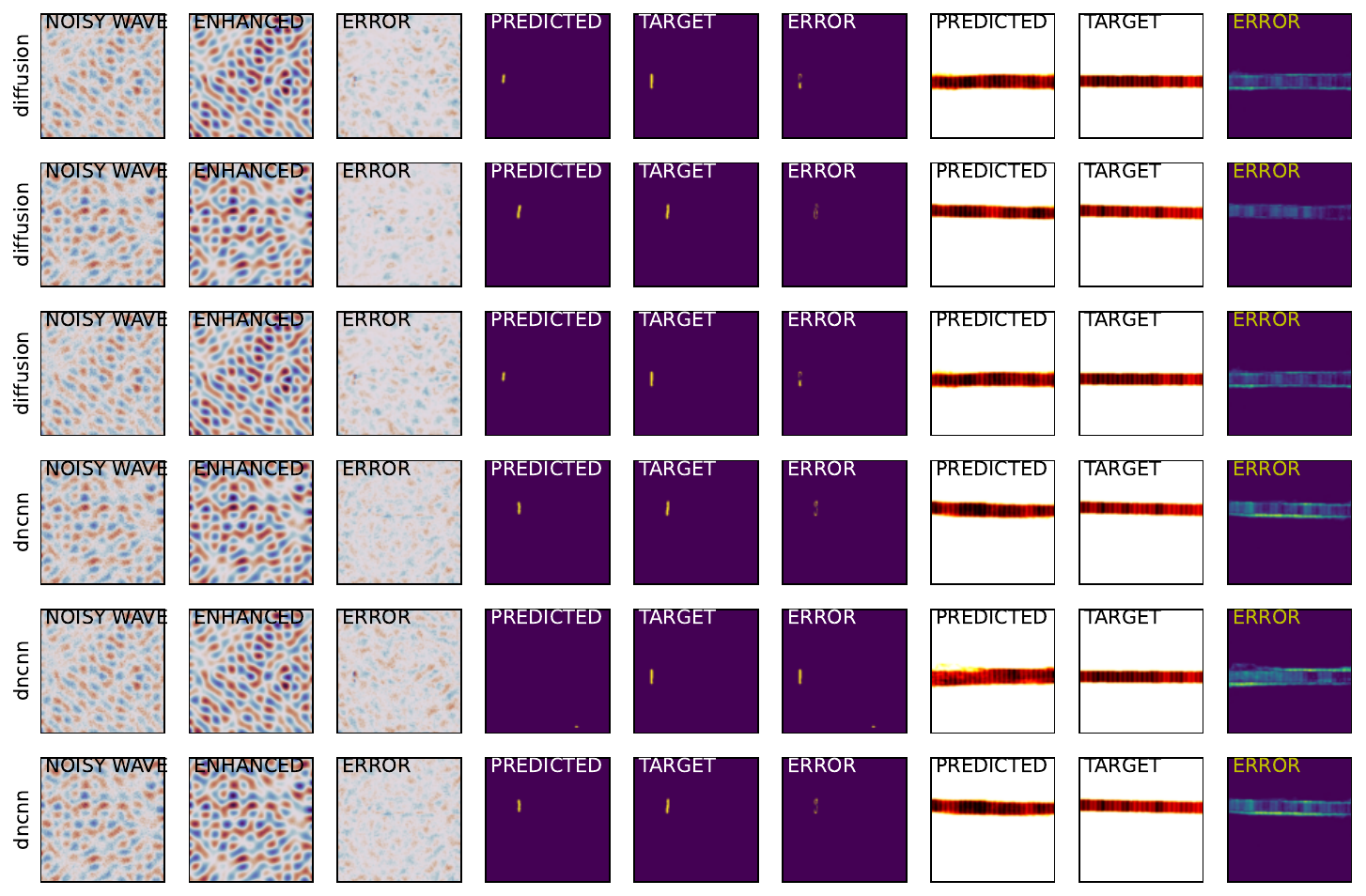}
	\caption{Performance comparison on synthetically-noised simulated data processed via conditional DDPM and DnCNN denoising.}
	\label{FIG:DnCNN_v_DDPM_Sim}
\end{figure}

\begin{figure}[t!]
	\begin{subfigure}{.5\linewidth}
		\includegraphics[width=\linewidth]{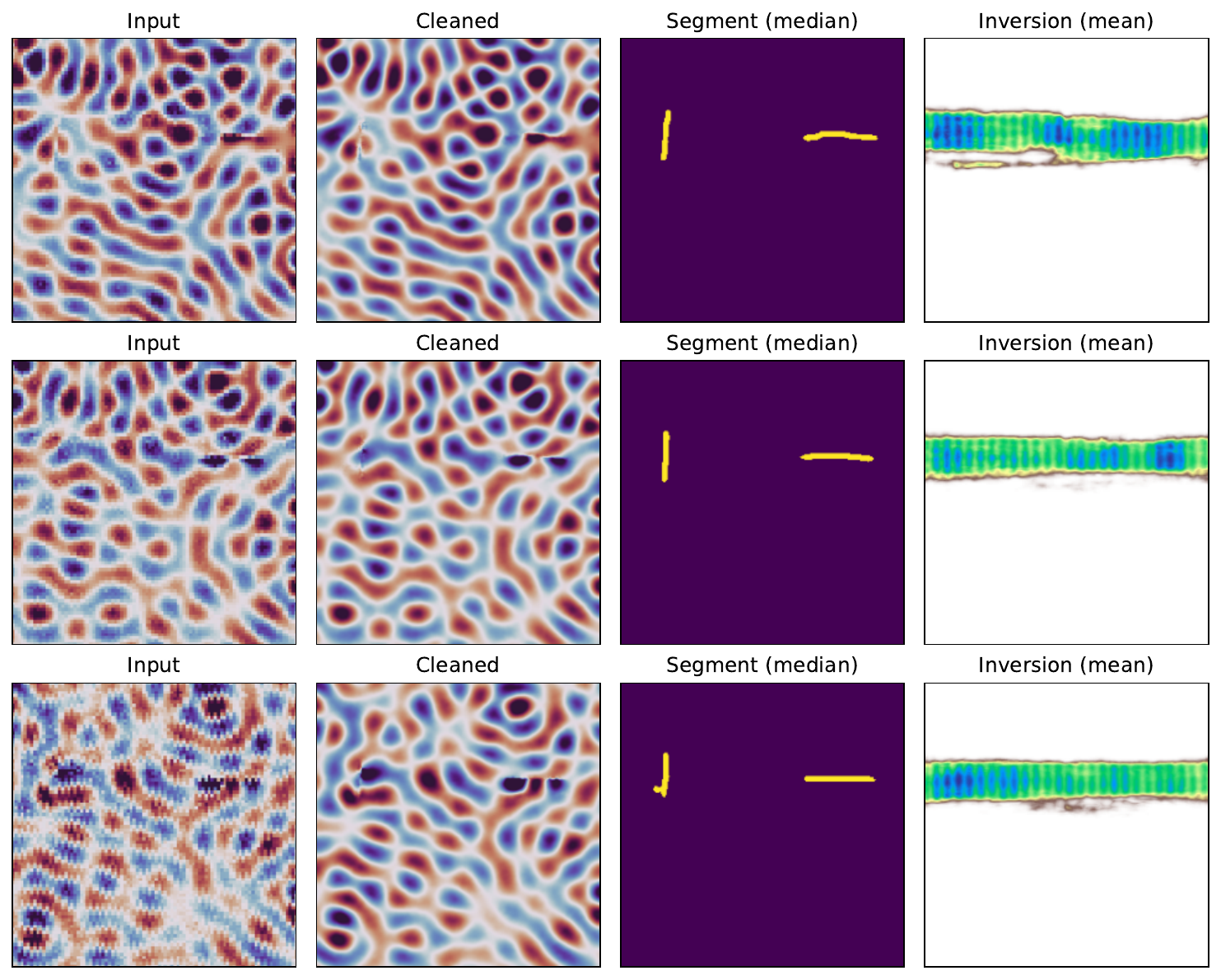}
		\caption{DDPM, No AWGN}
	\end{subfigure}%
	\begin{subfigure}{.5\linewidth}
		\includegraphics[width=\linewidth]{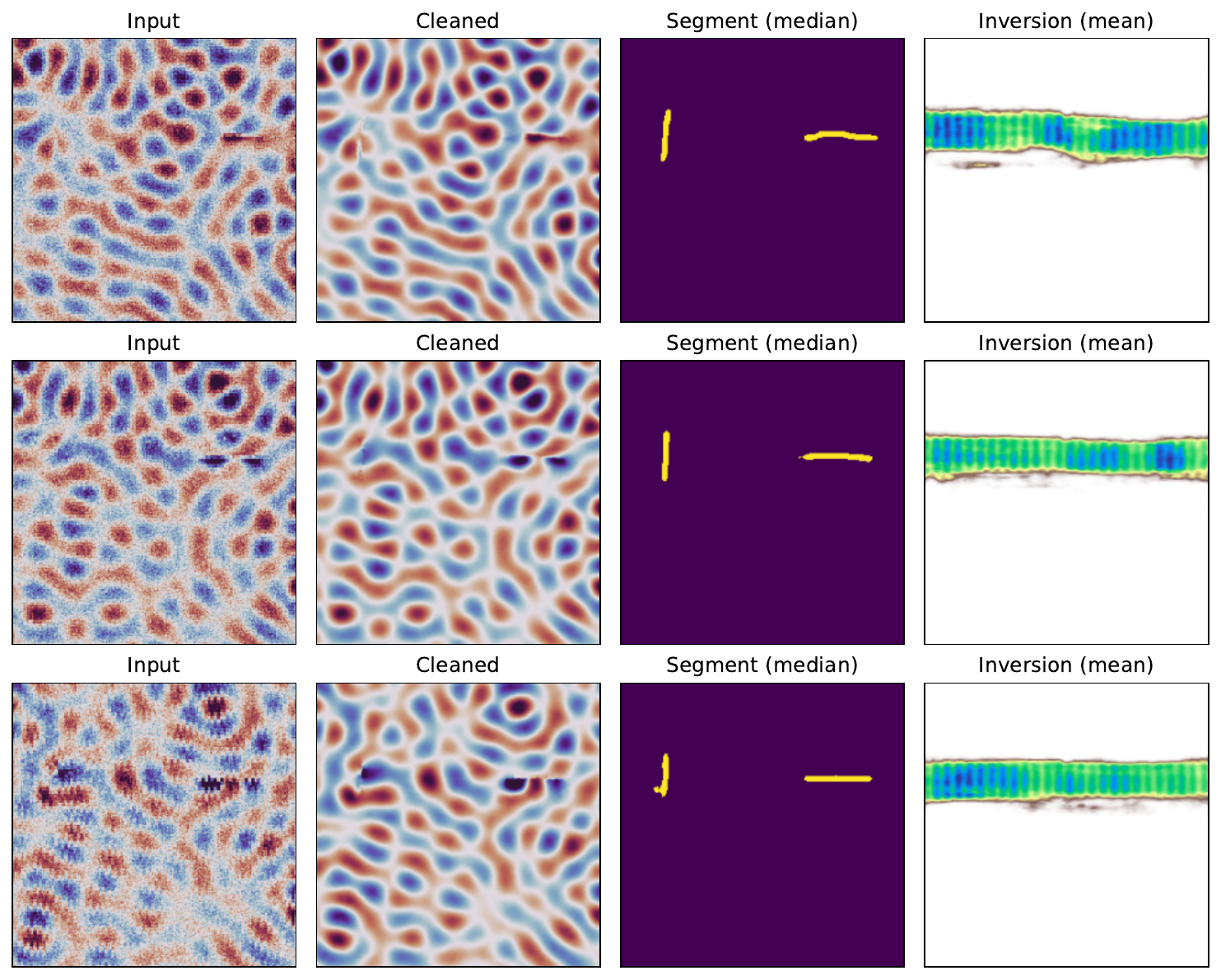}
		\caption{DDPM, 40\% AWGN}
	\end{subfigure}
	\begin{subfigure}{.5\linewidth}
		\includegraphics[width=\linewidth]{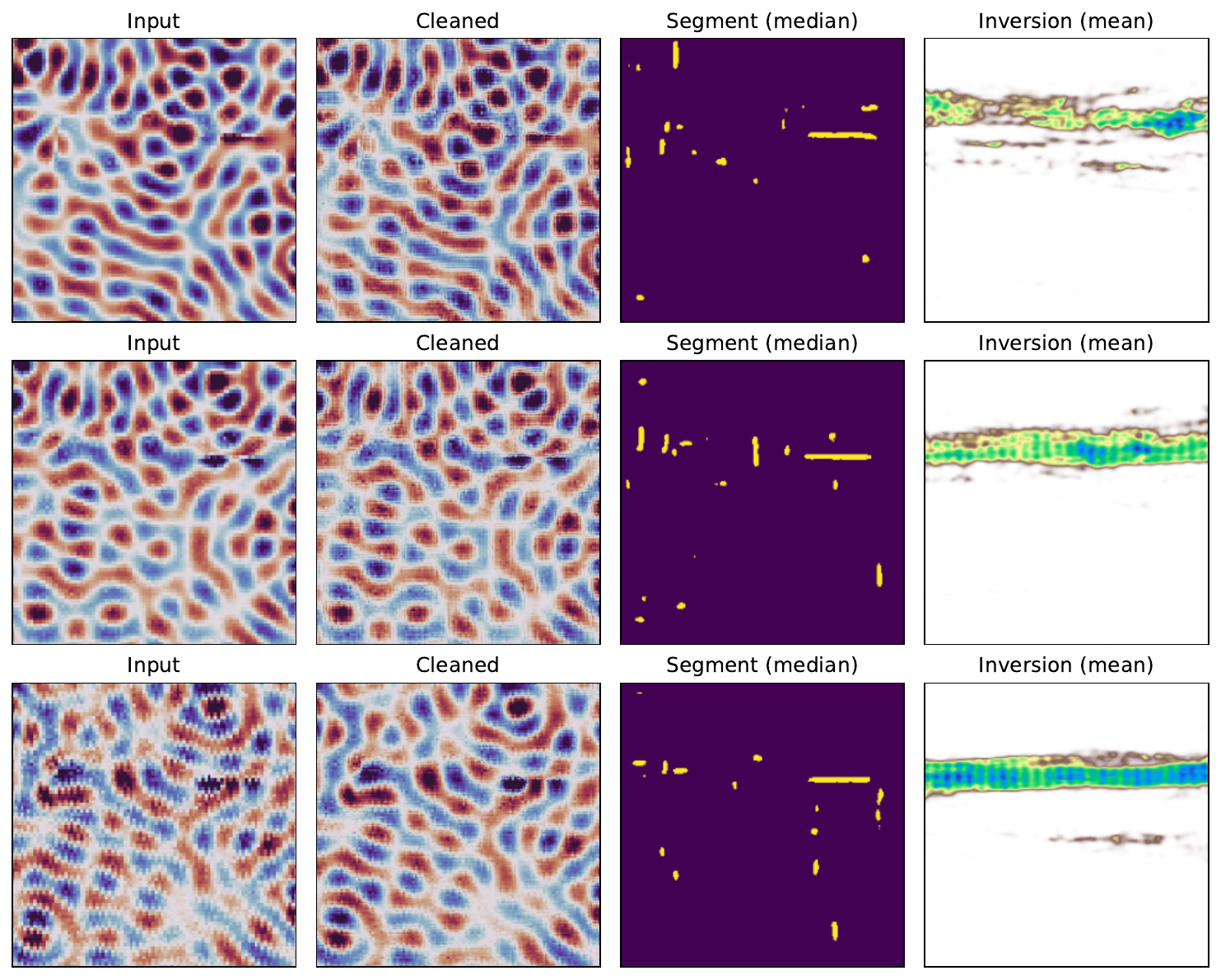}
		\caption{DnCNN, No AWGN}
	\end{subfigure}%
	\begin{subfigure}{.5\linewidth}
		\includegraphics[width=\linewidth]{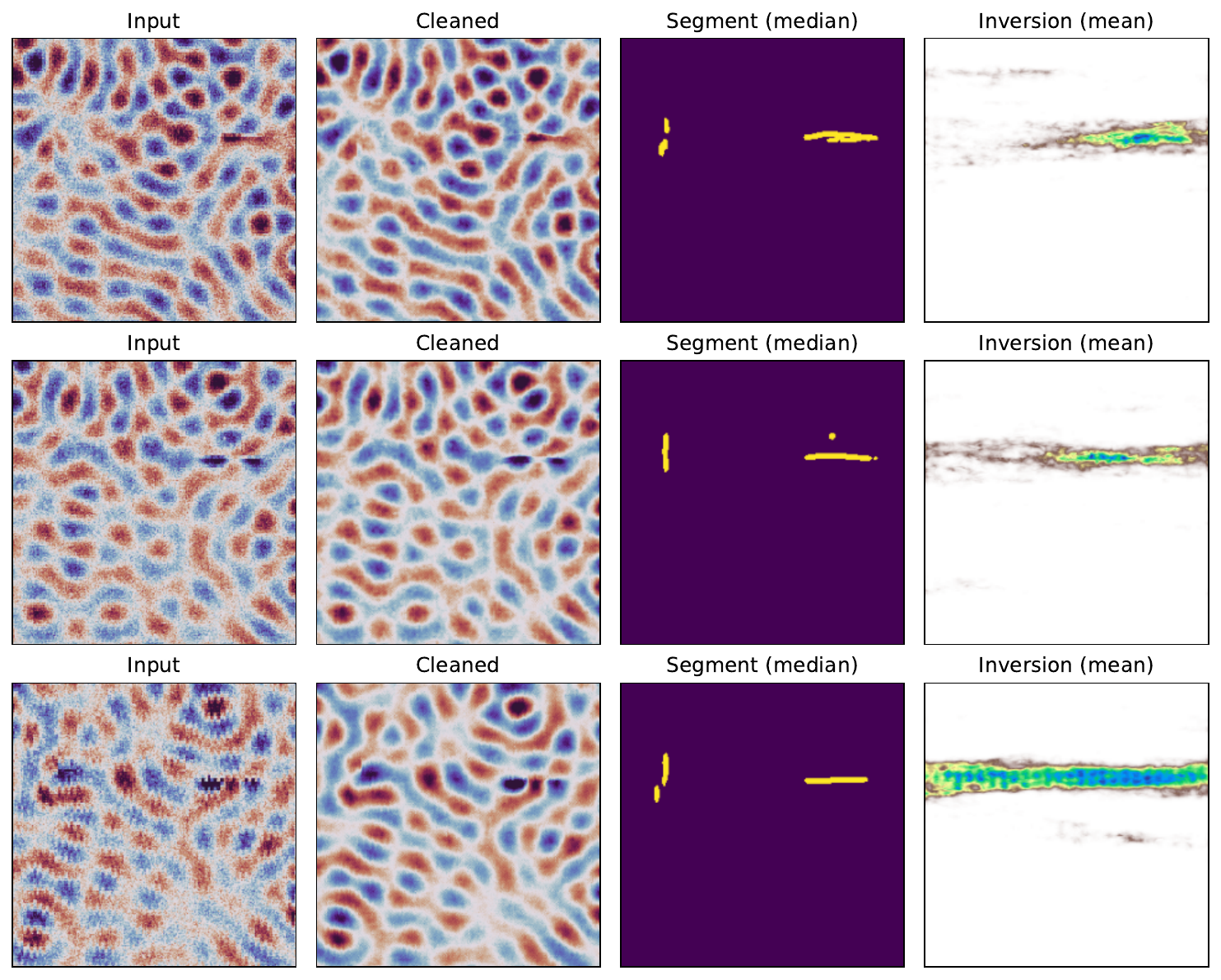}
		\caption{DnCNN, 40\% AWGN}
	\end{subfigure}
	\caption{Performance on ML-processed experimental measurements using (a,b) DDPM conditional generation and (c,d) DnCNN denoising, with (a,c) no additive white Gaussian noise (AWGN) and (b,d) 40\% AWGN.} 
	\label{FIG:DnCNN_v_DDPM_Exp}
\end{figure}

We compare the performance of our DDPM distribution shift model (cf. section~\ref{subsec:DDPM} and~\ref{subsubsec:DDPM_Guide}), we compare the performance against a DNCNN in both a numerical and experimental noise trial.
Fig~\ref{FIG:DnCNN_v_DDPM_Sim} depicts the results for three elastodynamic simulations of the coupon (free-free) problem class selected at random.
A predefined noise distribution taken as combination of Gaussian and speckle noise (i.e., Eq~\eqref{EQ:AddedNoise}) was added to the simulated wave to produce a \textit{noisy wave} (left column). Conditional generation as described in section~\ref{subsubsec:DDPM_Guide} rendered \textit{enhanced} representations of the noisy waves (first three rows), and a trained DNCNN was applied as a denoiser. The error in the resulting wavefields is low for both methods, and the subsequent predictions are also comparable (with a slight but not dominant performance improvement seen for the DDPM-based approach).  Thus, DNCNN works reasonably well when a predefined training noise distribution is known.

A comparison of the two methods on experimental data is given by Fig~\ref{FIG:DnCNN_v_DDPM_Exp}, where a clear advantage is now seen for the DDPM distribution-shift approach (as apposed to one-shot DNCNN denoising). 
Sub-panels (a) and (c) compare the input and cleaned wavfields as well as the model inference for DDPM and DNCNN, respectively. Sub-panels (b) and (d) show the same, but with 40\% added white Gaussian noise to the input measurement (this is done in an attempt to bring the measured waves closer to the training distribution of the DNCNN). 
In the former case, DNCNN produces wavefeilds that can not be meaningfully interpreted by the inverse models. In the latter, DNCNN performance is seemingly improved for just the segmentation task. 
In contrast, the DDPM distribution shift approach provides stable and meaningful results for all scenarios, emphasizing its effectiveness and robustness to data not captured in training.

\end{document}